\documentclass{ieeeaccess}
\usepackage{cite}
\usepackage{amsmath,amssymb,amsfonts}
\usepackage{algorithmic}
\usepackage{graphicx}
\usepackage{textcomp}
\usepackage{tabularx} % Make sure to include this package in your preamble

\usepackage{booktabs}
\usepackage[margin=1in]{geometry}
\usepackage{array}  % for column formatting
\usepackage{float}
\usepackage{placeins}
\usepackage{bm}
\usepackage{enumitem}   % For creating compact itemized lists inside cells
\usepackage{array}      % For advanced column specifications (like in the link column)
\usepackage[colorlinks=true, urlcolor=blue, linkcolor=black]{hyperref} % For clickable URLs
\usepackage{cleveref}

\makeatletter
\AtBeginDocument{\DeclareMathVersion{bold}
\SetSymbolFont{operators}{bold}{T1}{times}{b}{n}
\SetSymbolFont{NewLetters}{bold}{T1}{times}{b}{it}
\SetMathAlphabet{\mathrm}{bold}{T1}{times}{b}{n}
\SetMathAlphabet{\mathit}{bold}{T1}{times}{b}{it}
\SetMathAlphabet{\mathbf}{bold}{T1}{times}{b}{n}
\SetMathAlphabet{\mathtt}{bold}{OT1}{pcr}{b}{n}
\SetSymbolFont{symbols}{bold}{OMS}{cmsy}{b}{n}
\renewcommand\boldmath{\@nomath\boldmath\mathversion{bold}}}
\makeatother

\def\BibTeX{{\rm B\kern-.05em{\sc i\kern-.025em b}\kern-.08em
    T\kern-.1667em\lower.7ex\hbox{E}\kern-.125emX}}

\begin{document}
\history{Date of publication xxxx 00, 0000, date of current version xxxx 00, 0000.}
\doi{10.1109/ACCESS.2024.0429000}

\title{Learning by Watching: A Review of Video-based Learning Approaches for Robot Manipulation}
\author{\uppercase{Chrisantus Eze}\authorrefmark{1}, %\IEEEmembership{Fellow, IEEE},
\uppercase{Christopher Crick}\authorrefmark{2}}

\address[1,2]{Computer Science Department, Oklahoma State University, Stillwater, Oklahoma, USA}

\tfootnote{}

\markboth
{Chrisantus Eze \headeretal: Learning by Watching}
{Chrisantus Eze \headeretal: Learning by Watching}

\corresp{Corresponding author: Chrisantus Eze (e-mail: chrisantus.eze@okstate.edu).}

\begin{abstract}
Robot learning of manipulation skills is hindered by the scarcity of
diverse, unbiased datasets. While curated datasets can help,
challenges remain regarding generalizability and real-world
transfer. Meanwhile, large-scale 'in-the-wild' video datasets have
driven progress in computer vision using self-supervised
techniques. Translating this to robotics, recent works have explored
learning manipulation skills using abundant passive videos
sourced online. Showing promising results, such video-based learning
paradigms provide scalable supervision and reduce dataset
bias. This survey reviews foundations such as video feature
representation learning techniques, object affordance understanding,
3D hand and body modeling, and large-scale robotic resources, as well as
emerging techniques for acquiring robot manipulation skills from
uncontrolled video demonstrations. We discuss how learning from
only observing large-scale human videos can enhance generalization and
sample efficiency for robotic manipulation. The survey summarizes
video-based learning approaches, analyzes their benefits over standard
datasets, survey metrics and benchmarks, and discusses open
challenges and future directions in this nascent domain at the
intersection of computer vision, natural language processing, and
robot learning.
\end{abstract}


\titlepgskip=-21pt

\maketitle

\section{Introduction}
\label{sec:introduction}
\PARstart{I}{n} contrast to fields like computer vision (CV) and natural language processing (NLP), where copious amounts of high-quality and diverse datasets are available, the field of robotics faces a significant limitation in the availability of such datasets for various tasks. This scarcity of quality data has hindered progress in robotics in multiple ways. To address this challenge, researchers have proposed algorithms based on techniques like few-shot learning \cite{wang2020generalizing,james2018task,kadam2020review} and multitask learning \cite{finn2017one,yu2018one}. While these
approaches show promise in mitigating the data scarcity issue, they still rely on a substantial amount of high-quality data for effective task generalization.

Similarly, classical robot planning and manipulation methods often necessitate detailed modeling of the world and agent dynamics, further limiting their transferability and generalizability. Despite efforts to employ deep reinforcement learning (RL) for motion planning \cite{qureshi2020motion}, \cite{fishman2023motion} and manipulation \cite{rajeswaran2017learning}, \cite{zhu2019dexterous}, these methods encounter challenges such as distribution shifts and reductions in generalizability. Recent imitation learning methods \cite{jang2022bc,qin2022dexmv,shridhar2022cliport} based on Behavioral Cloning (BC) \cite{torabi2018behavioral} have also emerged as a potential solution to learning manipulation skills from minimal demonstrations. However, similar to their deep RL counterparts, these methods struggle to learn manipulation skills in diverse and uncurated datasets.

In recent times, significant strides have been made in collating large-scale, high-quality datasets for diverse robotic tasks \cite{brohan2022rt,zitkovich2023rt,padalkar2023open}, \cite{walke2023bridgedata,dasari2019robonet,fang2023rh20t} akin to the impact of the ImageNet dataset in the field of Computer Vision \cite{russakovsky2015imagenet}. While this marks a positive step forward, these datasets often exhibit limitations in their representativeness of real-world environments, as they are typically collected in controlled settings. Despite their advantages, these datasets pose potential drawbacks, including limited generalizability, biases \cite{gupta2018robot}, high costs, and ethical concerns regarding the interactions of embodied agents with humans.

In contrast, "in-the-wild" datasets have played a pivotal role in the success of computer vision \cite{goyal2021self,tian2021divide,caron2019unsupervised,miech2020end,cui2022play}, particularly with the rise of self-supervised learning. In the realm of robotics, various works have embraced this approach, training embodied agents to acquire manipulation skills by learning from videos sourced from platforms like YouTube. These endeavors have demonstrated impressive performance improvements, showcasing enhanced generalizability.

This paper provides a comprehensive exploration of video-based learning methodologies, with a focus on addressing fundamental challenges in vision-based robotic manipulation. Specifically, we investigate the potential of these methodologies to enhance the learning of generalizable skills, mitigate biases, and reduce the costs associated with curating high-quality datasets. Our contributions are threefold: (1) a detailed review and analysis of the capabilities of current approaches in various robotic tasks, (2) an overview of some open-source resources and tools for video-based robot manipulation learning to help researchers get started, and (3) a discussion of current challenges and future directions in the field. Our work focuses solely on vision-based manipulation. We discuss navigation, locomotion, and non-visual-based approaches only to provide a more broad perspective.  We commence by introducing and summarizing the foundational components of learning from videos, and then proceed to discuss current approaches for acquiring manipulation skills through video-based learning.

In Section \ref{sec:foundation}, we delineate and discuss the pipeline and essential components required for learning from video data. Additionally, we present notable large-scale robotic resources, including datasets and network architectures. Section \ref{sec:approaches} delves into the current approaches for learning from videos, categorized into five distinct groups, with a thorough literature review under each category. Section \ref{sec:comparative_analysis} highlights the comparative analysis of the distinct categories of approaches. We present in Section \ref{sec:open_source_tools} an overview of some open-source resources and tools used for video-based manipulation skill learning. Finally, Sections
\ref{sec:challenges} and \ref{sec:future_outlook} summarize the
existing challenges faced by researchers in developing systems for
learning manipulation skills from videos, and propose potential
research directions likely to have a significant impact in this
domain.

\subsection{Scope of this Survey}
\label{sec:scope}
This survey specifically explores techniques for acquiring robot manipulation skills through explicit learning from video data. Discussions on learning manipulation skills from data modalities other than videos or topics related to learning robot navigation skills are not within the scope of this article, though they may be briefly mentioned for a more comprehensive perspective. While foundational resources supporting this learning approach are touched upon, the primary focus is on discussing the techniques, advantages, and challenges associated with acquiring robust manipulation skills from video data. It is important to clarify that the list of foundational resources provided is not exhaustive and represents only the essential secondary components involved in the learning process. To the best of our knowledge, this survey is the first of its kind to explore the landscape of learning robot manipulation skills specifically from video data. Additionally, there is currently no existing work that comprehensively surveys the learning of robot skills in general, from videos, although many studies have surveyed the broader fields of robot manipulation and robot learning from demonstration.

\subsection{Related Surveys}
\label{sec:related_surveys}
We examine survey articles already available on related subjects to
guide readers to additional papers focusing on more specific
topics. This serves the dual purpose of offering references for
further exploration and elucidating the distinctions between this
article and existing surveys.

In contrast to these prior surveys highlighted in Table \ref{tab:related_survey}, our work is uniquely focused on the intersection of video-based learning and robot manipulation. We systematically review and compare methods across both imitation and reinforcement learning that leverage video demonstrations as a core component, encompassing advances in vision-language models (VLMs), foundation models, and large-scale data resources. By specifically addressing challenges, generalization, and open questions in video-based manipulation learning, our survey fills a notable gap in the literature and provides a valuable reference for researchers seeking to navigate this rapidly evolving subfield.

We summarize and compare the specific areas of focus of these survey articles with ours in Table \ref{tab:related_survey} below.

\begin{table*}[htbp]
    \centering
    \small
    \setlength{\tabcolsep}{4pt}
    \begin{tabular}{c c c c c c }
        \toprule
        \textbf{Paper} & \textbf{Cluttered Envs} & \textbf{LfD Techniques} & \textbf{RL} & \textbf{Generalized Algorithms} & \textbf{Foundation Models} \\
        \hline
        \cite{s22207938} & \checkmark & x & x & \checkmark & x \\
        \cite{argall2009survey} & x & \checkmark & x & x & x \\
        \cite{ravichandar2020recent} & x & \checkmark & x & x & x \\
        \cite{robotics10030105} & x & x & \checkmark & \checkmark & x \\
        \cite{kroemer2021review} & x & x & \checkmark & x & x \\
        \cite{xiao2023robot} & x & x & x & x & \checkmark \\
        \cite{zeng2023large} & x & x & x & x & \checkmark \\
        \cite{mccarthy2024towards} & x & x & \checkmark & x & \checkmark \\
        Ours & \checkmark & \checkmark & \checkmark & x & \checkmark \\
        \bottomrule
    \end{tabular}
\caption{A comparison of different aspects covered by existing surveys
  and our survey}
\label{tab:related_survey}
\end{table*}

\section{Foundations of Learning from Videos}
\label{sec:foundation}
Learning robot manipulation skills from videos is a complex task that
necessitates a comprehensive visual pipeline, encompassing various
objectives such as representation learning, object affordance
learning, human action recognition, and 3D hand modeling. In this
section, we will delve into these objectives in detail and explore
some of the proposed techniques for their execution.

\subsection{Representation Learning}
\label{sec:rep_learning_foundation}

Visual feature extraction forms the backbone of vision-based robotics. Over time diverse representation learning methods have emerged, each offering unique ways to capture meaningful features from visual data. These approaches broadly fall into two categories: those tailored for videos and those applicable to both images and videos.

Video-centric representation learning focuses on modeling temporal dynamics and multi-view consistency. For instance, Time-Contrastive Networks (TCN) \cite{sermanet2018time} use self-supervised learning from multi-view videos to encode temporal changes while remaining invariant to viewpoint differences. Building on this idea, Domain-agnostic Video Discriminator (DVD) \cite{chen2021learning} employs multitask reward learning, training a discriminator to verify whether two videos depict the same task, thereby extracting domain-invariant features.

Unsupervised approaches further extend these capabilities. Wang et al. \cite{wang2015unsupervised} leverage visual tracking as supervision, aligning tracked patches across frames via a Siamese-triplet loss to learn rich representations from unlabelled web videos. While CNNs excel at spatial feature extraction, temporal modeling often benefits from sequence models. For example, \cite{srivastava2015unsupervised} introduced an LSTM-based encoder-decoder for compact video representations, supporting sequence reconstruction and future prediction. Similarly, Dense Predictive Coding (DPC) \cite{han2019video} learns spatio-temporal embeddings in a self-supervised manner for tasks like action recognition.

Beyond spatial and temporal cues, some methods incorporate geometry and structure. \cite{zhou2017unsupervised} proposed an unsupervised framework for jointly estimating monocular depth and camera motion using view synthesis as supervision, inspiring later work on 3D scene understanding \cite{flynn2016deepstereo,garg2016unsupervised,godard2017unsupervised}. Other methods, such as Contrastive Video Representation Learning (CVRL) \cite{qian2021spatiotemporal}, use contrastive learning to align augmented views while differentiating unrelated clips, producing robust spatiotemporal representations.

Moving toward general-purpose methods, recent research targets representations transferable across images and videos. Masked Modeling \cite{xiao2022masked} showed that self-supervised pretraining on real-world images can outperform traditional ImageNet-based pretraining \cite{russakovsky2015imagenet} in robotic manipulation benchmarks. Extending this idea, \cite{radosavovic2023real} applied masked autoencoders (MAE) to large-scale video data for visual pretraining, integrating these frozen representations into downstream control policies.

Finally, universal representations like R3M \cite{nair2022r3m} combine time-contrastive learning with sparse encoding from human video datasets. Serving as a frozen perception module, R3M enables efficient imitation learning across both simulated and real-world robotic manipulation tasks.

Table \ref{tab:quantitative_comparison} shows a concise summary of the comparison between the approaches discussed.

\begin{table*}[h!]
\centering
\footnotesize
\begin{tabular}{p{1.1cm} p{3.0cm} p{3.0cm} p{2.5cm} p{3.2cm}}
\toprule
\textbf{Method} & \textbf{Task Performance} & \textbf{Sample Efficiency} & \textbf{Generalization} & \textbf{Compute Cost} \\
\midrule

\cite{sermanet2018time} & Effective for third-person imitation; supports RL via embedding similarity & Learns from multi-view video; label-free but needs synchronization & Generalizes across views and agents with consistent context & Lightweight CNN + metric loss; training setup moderately complex \\
\midrule

\cite{chen2021learning} & Solves real-robot tasks using learned video rewards & Requires only a few robot demos and 1 human video & Strong zero-shot generalization to new tasks and environments & Moderate; uses discriminators with broad human video data \\
\midrule

\cite{wang2015unsupervised} & Nearly matches ImageNet CNNs on VOC (52\% mAP) & Uses 100K videos and patch tracking; no labels needed & Captures object-level similarity; weak on temporal cues & Siamese-triplet CNN; efficient but needs large-scale video processing \\
\midrule

\cite{srivastava2015unsupervised} & Predicts future video frames; helps with action recognition & Requires many sequences; less data-efficient & Weak on domain shift; struggles with unfamiliar dynamics & Moderate; encoder-decoder LSTMs for temporal modeling \\
\midrule

\cite{han2019video} & 75.7\% on UCF101; strong for human action recognition & Avoids pixel prediction; uses curriculum for better efficiency & Robust to viewpoint and appearance variation & 3D-ResNet + GRU; heavier than 2D CNNs but no reconstruction loss \\
\midrule

\cite{zhou2017unsupervised} & Effective depth and ego-motion learning from video & Self-supervised via view synthesis; highly data-efficient & Generalizes across driving scenes; handles occlusions & Dual CNNs for depth and pose; warping increases training cost \\
\midrule

\cite{qian2021spatiotemporal} & 70.4\% on Kinetics-600; outperforms SimCLR and ImageNet & Leverages contrastive loss with smart augmentations & Temporal and spatial robustness; handles distant clip variations & R3D-50 backbone; moderate complexity with consistent training speed \\
\midrule

\cite{xiao2022masked} & Solves motor control tasks; up to 80\% success & Pretrained on real-world images; no labels or robot data needed & Generalizes across tasks and robots & ViT-based MAE; costly pretraining, efficient inference \\
\midrule

\cite{radosavovic2023real} & 81\% success in real-world robot tasks & Achieves strong results with only 20-80 demos per task & Generalizes across tasks, robots, and scenes & 307M ViT encoder; high initial cost, efficient during deployment \\
\midrule

\cite{nair2022r3m} & +20\% over MoCo/CLIP on 12 tasks; 50\%+ success from 20 demos & Learns from human videos; effective with few demos & Strong task, view, and embodiment transfer & Sparse contrastive encoder; efficient for downstream control \\

\bottomrule
\end{tabular}
\caption{Comparison of representative video-based visual representation learning methods for robot manipulation. }
\label{tab:quantitative_comparison}
\end{table*}

\subsection{Object Affordance and Human-Object Interaction}
\label{sec:object_affordance}

A key step in enabling robots to acquire manipulation skills from videos is understanding object affordances: the actionable properties of objects, through the lens of human interaction. Over the years, research has progressed from early hand-state analysis in large-scale internet videos to increasingly sophisticated, multimodal, and context-aware frameworks. This subsection charts the evolution of these methods, highlighting their innovations and interconnections.

We begin with approaches that leverage large-scale, unstructured human activity videos to uncover affordances. For example, \cite{shan2020understanding} extracts hand-state information from internet videos, laying a foundation for understanding human-object interaction at scale. Building on this,  Hand-aided Affordance Grounding Network (HAG-Net) \cite{luo2023learning} employs hand cues from demonstration videos and a dual-branch network for fine-grained localization of affordance regions, improving the precision of affordance grounding.

The field then expands toward capturing functional understanding and temporal dynamics of affordances. The authors in \cite{koppula2014physically} introduce a generative model that grounds object affordances by considering both spatial context and human intention, while a related approach \cite{koppula2013learning} models objects and sub-activities as a Markov random field, addressing the challenge of acquiring descriptive labels for sub-activities and their corresponding affordances.

Learning from demonstration (LfD) videos is another significant direction. Demo2Vec \cite{fang2018demo2vec} focuses on reasoning about object affordances using carefully curated demonstration videos, learning vector embeddings to predict interaction regions and support both human and robot understanding. In contrast, Vision-Robotics Bridge (VRB) \cite{bahl2023affordances} demonstrates how affordance models trained on diverse, in-the-wild internet videos can bridge the gap between human-centric video data and the requirements of robotic applications.

The integration of depth data and multi-modal inputs has further advanced affordance detection. AffordanceNet \cite{do2018affordancenet} exemplifies this by introducing an end-to-end deep learning method for identifying both objects and their affordances in RGB-D images, effectively handling multiclass affordance masks. Similarly, \cite{williams2019analysis} reviewed the landscape of affordance detection methods, underscoring the importance of understanding the full range of object affordances for real-world robot intelligence. At the 3D level, \cite{kim2014semantic} proposed semantic labeling of 3D point clouds to improve object segmentation and reduce uncertainty in manipulation, demonstrating that incremental, multi-view merging can directly benefit manipulation planning.

The scope broadens to include human-object relationships and interaction recognition in video. The authors in \cite{ji2021detecting} employed transformer architectures for joint spatial-temporal reasoning, while \cite{goyal2022human} leveraged hand localization in egocentric videos to understand object affordances. H20 dataset was introduced in \cite{kwon2021h2o}, enabling synchronized multi-view RGB-D capture of two-handed object manipulation, providing rich annotations for developing and benchmarking new affordance-centric methods.

Moving from interaction recognition to grasp generation and segmentation, \cite{jiang2021hand} highlights the consistency between hand contact points and object regions, introducing objectives for self-supervised training of grasp generation models. \cite{tekin2019h+} proposed a unified network to simultaneously predict 3D hand and object poses, model interactions, and recognize action categories in egocentric video sequences. Meanwhile, \cite{shan2021cohesiv} developed a weakly supervised approach to segment hands and hand-held objects from motion in a single RGB image, leveraging motion-derived responsibility maps for network training.

Datasets play a pivotal role in advancing affordance learning. The authors in \cite{garcia2018first} introduced a comprehensive dataset with over 100K frames of hand-object interactions and rich 3D annotations. In parallel, \cite{cai2016understanding} used computer vision to unify the identification of hand grip types, object properties, and action categories from images, providing a context-aware model of natural hand-object manipulation.

\subsection{Human Action and Activity Recognition}
\label{sec:human_action}

Recognizing human actions is essential for robots operating in human environments, particularly for understanding object affordances. Early work leveraged human action patterns from videos to guide robotic perception and decision-making \cite{luo2023learning,koppula2013learning,shan2020understanding,koppula2015anticipating}. Building on this, \cite{pieropan2013functional} emphasized affordances over appearance, encoding object-hand interactions as strings to capture functional properties more robustly. To enable fine-grained recognition, \cite{ma2018region} combined appearance and motion cues through convolutional networks, enhancing the discrimination of subtle action variations. For automated object interaction analysis, Interaction Region and Motion Trajectory prediction Network (IRMT-Net) \cite{xin2023learning} jointly estimates interaction regions and motion trajectories from demonstrations, reducing reliance on manual guidance and improving adaptability across systems.

Recent advances include unsupervised and weakly supervised methods. \cite{sener2018unsupervised} proposed a framework that segments actions into sub-activities using alternating discriminative and generative learning, coupled with background modeling to filter irrelevant frames, achieving strong performance with minimal labeled data. Progress has also been driven by large-scale datasets such as UCF101 \cite{soomro2012ucf101}, which provides diverse, realistic user-uploaded videos for benchmarking action recognition algorithms.

\subsection{3D Hand Modeling}
\label{sec:3d_hand}

Bridging the embodiment gap between human hands and robot grippers is a key challenge in learning manipulation skills from videos. To address this, researchers have explored both hardware solutions and computational models for 2D/3D hand representation, enabling robots to more effectively imitate human actions.

Early approaches introduced anthropomorphic robotic hands for teleoperation and video-based learning. For example, LEAP Hand \cite{shaw2023leap} is a low-cost design that supports visual teleoperation, passive learning, and sim-to-real transfer by extracting hand poses from web videos. Similarly, DexMV \cite{qin2022dexmv} provides a simulation and vision-based pipeline that maps 3D human hand poses to robot-compatible demonstrations, while DexVIP \cite{mandikal2022dexvip} leverages YouTube videos and human hand priors to learn dexterous grasping without expensive lab data, enabling generalization to novel objects.

Advances in motion capture have further improved fidelity. FrankMocap \cite{rong2020frankmocap} offers fast monocular 3D hand and body pose estimation using SMPL-X \cite{SMPL-X:2019}, while MANO \cite{romero2022embodied} provides a parametric hand model built on SMPL \cite{SMPL:2015}, delivering low-dimensional, realistic representations widely used in robotics and graphics. Complementary work \cite{jiang2021hand} focuses on grasp realism, enforcing consistency between hand-object contact points through self-supervised objectives, improving flexibility and accuracy even during testing.

Collectively, these methods reduce embodiment differences and establish robust hand modeling pipelines, laying the foundation for high-fidelity manipulation learning from human demonstrations.

\subsection{Datasets}
\label{sec:datasets}
In our discussion of the essential components for training robot manipulation policies, the importance of datasets cannot be overstated. Datasets form the foundation upon which learning algorithms build their understanding of manipulation tasks. Learning
robot manipulation skills from demonstration videos requires carefully curated datasets that capture humans performing these tasks in various environments such as kitchens, living rooms, workshops, and more. These datasets not only provide the raw data necessary for training but also offer insights into human-object interactions, task variability, and environmental context.

This section categorizes and details some of the most influential
datasets used in the field of robot manipulation learning,
highlighting their unique characteristics and contributions.

\subsubsection{Large-Scale Video Datasets}
These datasets offer a vast amount of video data capturing diverse activities and interactions. Additionally, these datasets are typically sourced from the internet and present a wide range of scenarios and tasks, making them invaluable for generalizing robot learning.

\begin{itemize}
\item \textbf{YouTube}: As one of the largest video platforms, YouTube serves
  as a rich source of diverse video content. Several works have
  curated specific subsets of YouTube videos relevant to robots
  manipulation, providing a broad spectrum of tasks and environments.
  
  For instance, in \cite{xue2022advancing}, the authors introduced
  HD-VILA-100M, a large dataset with two distinct properties: 1) it is
  the first high-resolution dataset, including 371.5k hours of 720p
  videos, and 2) it is the most diversified dataset, covering 15
  popular YouTube categories. YT-Temporal-180M, introduced in
  \cite{zellers2021merlot}, is a diverse corpus of frames/ASR derived
  from a filtered set of 6M diverse YouTube videos.
    
  In addition to these works, researchers have introduced datasets
  with smoother and more descriptive video-text pairs. One of such
  works is WTS-70M, a 70M video clips dataset presented in
  \cite{stroud2020learning}, contains textual descriptions of the most
  important content in the video, such as the objects in the scene and
  the actions being performed. The authors in
  \cite{miech2019howto100m} introduced HowTo100M: a large-scale
  dataset of 136 million video clips sourced from 1.22M narrated
  instructional web videos depicting humans performing and describing
  over 23k different visual tasks. Additionally, \cite{bain2021frozen}
  provided a new video-text pretraining dataset WebVid-10M, comprised
  of over two million videos with weak captions scraped from the
  internet.
    
\item \textbf{Internvid}: Compiled from various internet sources, Internvid
  \cite{wang2023internvid} focuses on activities and tasks that are
  particularly informative for robotic learning. This dataset
  encompasses a wide array of human activities, enhancing the
  versatility of trained models.
    
\item \textbf{Something-Something}: This dataset \cite{goyal2017something}
  consists of videos where humans perform a wide range of actions on
  everyday objects. It is particularly useful for training models to
  recognize and replicate specific human-object interactions.
\end{itemize}

\subsubsection{Egocentric (First-Person) Video Datasets}
Egocentric datasets capture videos from the first-person perspective,
offering a unique vantage point for understanding hand-object
interactions and human intent. These datasets are especially valuable
for tasks that involve detailed manipulation and personal perspective.

\begin{itemize}
\item \textbf{Ego-4D}: A comprehensive dataset of first-person videos capturing
  daily activities, Ego-4D \cite{grauman2022ego4d} provides rich data
  on hand-object interactions from the wearer's perspective. This
  dataset is instrumental in training models to understand and predict
  human actions in a personal context.
    
\item \textbf{Ego-Exo-4D}: Building on the Ego-4D dataset, Ego-Exo-4D
  \cite{grauman2024ego} includes both egocentric and exocentric
  (third-person) views of the same activities. This multi-perspective
  approach offers a more holistic understanding of tasks, aiding in
  the development of models that can interpret and execute actions
  from different viewpoints.
\end{itemize}

\subsubsection{Task-Specific and Multi-Modal Datasets}
Task-specific and multi-modal datasets are designed to study particular tasks or provide multiple modalities of data, such as video, audio, and annotations. These datasets are tailored to enhance the learning process for specific manipulation skills.

\begin{itemize}
    \item \textbf{Epic Kitchens}: Focused on kitchen activities, this dataset \cite{damen2022rescaling} captures detailed interactions with objects and the environment from an egocentric perspective. The rich annotations and diversity of tasks make it ideal for training models on kitchen-related manipulation tasks.
    
    \item \textbf{RoboVQA}: This dataset \cite{sermanet2023robovqa} is designed for Visual Question Answering in robotic contexts. It includes videos of robots performing tasks and corresponding questions that test the robot's understanding and reasoning based on the visual data. RoboVQA helps in developing models that can interpret and respond to queries about manipulation tasks.
\end{itemize}

\subsubsection{Embodied AI and Interactive Datasets}
Embodied AI and interactive datasets emphasize tasks that involve
interaction with the environment, providing rich contextual
information that is crucial for learning manipulation skills.

\begin{itemize}
\item \textbf{Open X-Embodiment}: A comprehensive dataset
  \cite{padalkar2023open} that includes videos of various embodied AI
  tasks, capturing interactions in different environments. This dataset is the largest and most diverse open source robotics dataset to date, unifying 34 distinct datasets from 22 different robot embodiments. It is designed for large-scale, cross-platform model pretraining containing over 1.6 million trajectories spanning more than 60,000 unique tasks. It supports vision, language (instructions), and action (State/Pose) modalities.

\item \textbf{DROID (Distributed Robot Interaction Dataset)}:
DROID \cite{khazatsky2024droid} comprises over 76,000 trajectories (\textasciitilde{350} hours) collected across 564 scenes and 86 tasks by more than 50 users. Designed for diversity and generalization, it outperforms Open X-Embodiment on both in- and out-of-distribution tasks, making it a strong benchmark for imitation learning.

\item \textbf{BRMData (Bimanual-Mobile Robot Manipulation Dataset)}:
BRMData \cite{zhang2024empowering} is a dataset that focuses on dual-arm and mobile manipulation, capturing tasks such as object handovers, opening cabinets, and cleaning. It features ten household tasks performed by a mobile manipulator equipped with two robot arms, and includes RGB and depth data from multiple camera viewpoints. The dataset also emphasizes environmental interaction and whole-body planning, supporting the development of controllers that operate in both tabletop and mobile contexts.

\item \textbf{Fourier ActionNet}:
Fourier ActionNet \cite{fourier2025actionnet} contains 30,000 teleoperated bimanual trajectories (\textasciitilde{140} hours) of tabletop manipulation. Each trajectory is annotated with human-written task prompts, supporting instruction-conditioned policy learning and dexterous control.

\item \textbf{Kaiwu Dataset}:
Kaiwu \cite{jiang2025kaiwu} offers 11,664 demonstrations of human assembly tasks using 30 objects and 20 participants. The dataset includes synchronized RGB video, audio, EMG, eye gaze, motion capture, and tactile data, making it suitable for multimodal representation learning.

\item \textbf{TASTE-Rob}:
TASTE-Rob \cite{zhao2025taste} provides over 100,000 egocentric video clips of human manipulation aligned with natural language instructions. It emphasizes object-centric motion and temporal segmentation, useful for training video-conditioned imitation policies.

\end{itemize}

\begin{table*}[h!]
\centering
\footnotesize
\begin{tabularx}{\textwidth}{p{2.1cm} p{3.2cm} X p{2.8cm} p{2.5cm}}
\toprule
\textbf{Dataset} & \textbf{Content Focus} & \textbf{Key Feature} & \textbf{Scale} & \textbf{Modalities} \\
\midrule
\multicolumn{5}{l}{\textbf{General Web and Instructional Videos}} \\
\midrule
HowTo100M \cite{miech2019howto100m} & Instructional videos covering \textasciitilde{23}k different human tasks & Massive scale for learning procedural, step-by-step tasks & 136M clips from 1.22M videos (1.36M hours) & Vision, Language (ASR) \\
\addlinespace
WebVid-10M \cite{bain2021frozen} & Short, general-domain web videos with alt-text captions & Tightly-aligned, descriptive video-text pairs from web data & 10.7M clips from 2.5M videos (52k hours) & Vision, Language (Alt-text) \\
\addlinespace
Something-Something-v2 \cite{goyal2017something} & Basic human-object interactions (e.g., "pushing something") & Focus on fine-grained action recognition from templated labels & 220,847 video clips & Vision, Language (Labels) \\
\midrule
\multicolumn{5}{l}{\textbf{Egocentric Videos}} \\
\midrule
Ego-4D \cite{grauman2022ego4d} & Daily life activities captured from a first-person view & Unprecedented scale and diversity for egocentric human activity & 3,670 hours of video (2.78M clips) & Vision, Audio, Language, 3D Mesh, Eye Gaze, Stereo \\
\addlinespace
Ego-Exo-4D \cite{grauman2024ego} & Activities recorded simultaneously from ego- and exo-centric views & Provides paired perspectives for learning view-invariant skills & 1,400 hours of paired video & Vision, Audio, Language, Pose, 3D Geometry \\
\addlinespace
EPIC-KITCHENS-100 \cite{damen2022rescaling} & Unscripted activities in a kitchen environment & Dense, fine-grained annotations of actions and object interactions & 100 hours of video (90k action segments) & Vision, Audio, Language (Action/Object labels) \\
\midrule
\multicolumn{5}{l}{\textbf{Robotics and Embodied AI}} \\
\midrule
Open X-Embodiment \cite{padalkar2023open} & Manipulation trajectories from 22 different robot platforms & Unifies dozens of robotics datasets for large-scale co-training & 1M+ trajectories across 527 skills & Vision, Language (Instructions), Action (State/Pose) \\
\addlinespace
InternVid \cite{wang2023internvid} & Narrated videos of diverse human-object interactions & High-quality, cleaned video-text pairs for strong generalization & 236M clips from 7M videos (760k hours) & Vision, Language (Cleaned ASR) \\
\addlinespace
RoboVQA \cite{sermanet2023robovqa} & Robot manipulation sequences for visual reasoning tasks & Designed for question-answering about robot actions and states & 98k video-question pairs & Vision, Language (Q\&A) \\
\addlinespace
DROID \cite{khazatsky2024droid} & Diverse manipulation demonstrations across many scenes and tasks & Collected by 50+ users globally with high generalization capability & 76k trajectories (350 hours) & Vision, Language, Actions \\
\addlinespace
BRMData \cite{zhang2024empowering} & Mobile and dual-arm household manipulation & Captures both tabletop and mobile dual-arm tasks in real homes & 10 tasks with multi-view videos & Vision (RGB, Depth), Actions \\
\addlinespace
Fourier ActionNet \cite{fourier2025actionnet} & Bimanual dexterous manipulation & Teleoperated control with natural language task descriptions & 30k trajectories (140 hours) & Vision, Language (Prompts), Actions \\
\addlinespace
Kaiwu \cite{jiang2025kaiwu} & Human demonstrations of assembly tasks & Rich multimodal data including audio, gaze, EMG, and tactile sensors & 11,664 demos across 30 objects & Vision, Audio, EMG, Eye Gaze, Tactile, Motion Capture \\
\addlinespace
TASTE-Rob \cite{zhao2025taste} & Egocentric hand-object manipulation & Large-scale video-instruction pairs for manipulation learning & 100k+ video-instruction clips & Vision, Language (Instructions) \\
\bottomrule
\end{tabularx}
\caption{An overview of prominent video datasets relevant to robot learning. The table compares datasets on their primary content, unique features, scale, annotation methods, and data modalities.}
\label{tab:datasets_final}
\end{table*}

\subsection{Large Scale Robotic Resources}
\label{sec:large_resources}

The success of large-scale models in computer vision and natural language processing has set a high bar for what is possible with extensive data and powerful architectures. In robotics, a similar paradigm shift is underway with the introduction of Vision-Language-Action (VLA) models. The initial efforts in this space, represented by models like RT-1 \cite{brohan2022rt} and RT-2 \cite{zitkovich2023rt}, provided the foundational blueprint for large-scale robot learning.

RT-1 was a major milestone, introducing open-ended, task-agnostic training and the Robotics Transformer architecture to enable strong generalization to new tasks with minimal data. The model uses a FiLM-conditioned \cite{perez2018film} EfficientNet-B3 encoder, with instruction embeddings from a Universal Sentence Encoder, and compresses convolutional outputs using a TokenLearner module to produce compact visual tokens. These tokens, concatenated across the observation dimension, are fed into an 8-layer decoder-only transformer (\textasciitilde{19}M parameters) that autoregressively predicts discretized action tokens, with each action dimension quantized into 256 bins. This design enabled real-time control and established the feasibility of training open-ended, task-agnostic robot policies with strong generalization. Building on this, RT-2 advanced the field by integrating VLMs. Instead of training a small transformer from scratch, RT-2 used pretrained VLMs such as PaLI-X \cite{chen2023pali} and PaLM-E \cite{driess2023palm} as perception backbones, casting robot actions as discrete language tokens. Actions (e.g., 7-DOF poses and gripper states) were serialized into integer strings, enabling the model to predict them like words in a sentence. These models, along with collaborative efforts like Open X-Embodiment \cite{padalkar2023open}, established the viability of training adaptable policies on diverse datasets collected from multiple robot platforms. However, their reliance on discretization for action modeling imposed limitations in precision and multimodality. This laid the groundwork for the next wave of research, which seeks to overcome these constraints through architectural innovations in action representation and reasoning.

The limitations of first-generation large models have sparked a wave of models \cite{kim2024openvla,team2024octo,li2024cogact,liu2024rdt,cheang2024gr,bjorck2025gr00t,qu2025spatialvla,black2024pi_0} that prioritize more expressive action modeling and stronger reasoning capabilities. A central theme is the shift from simple action discretization to architectures that can capture continuity, multimodality, and temporal correlation in control.

One line of work extends the original discretization approach but enriches it with stronger perception and scaling. OpenVLA \cite{kim2024openvla}, for example, employs dual vision encoders (SigLIP \cite{zhai2023sigmoid} and DINOv2 \cite{oquab2023dinov2}) projected into a Llama-2 \cite{touvron2023llama} backbone, enabling action prediction as 256-bin tokens. Despite using just 7B parameters, OpenVLA outperforms larger closed-source counterparts such as RT-2-X \cite{padalkar2023open} (55B), largely due to training on nearly one million robot trajectories from the Open X-Embodiment corpus. SpatialVLA \cite{qu2025spatialvla} builds on this idea by embedding actions into “Adaptive Action Grids,” where motions are discretized into spatially grounded tokens tied to 3D coordinates. This spatially informed design, combined with Ego3D position encodings, improves sim-to-real transfer and allows fine-grained control across environments.

\begin{figure*}[htbp] % htbp stand for "here", "top", "bottom", "page"
\includegraphics[scale=0.33]{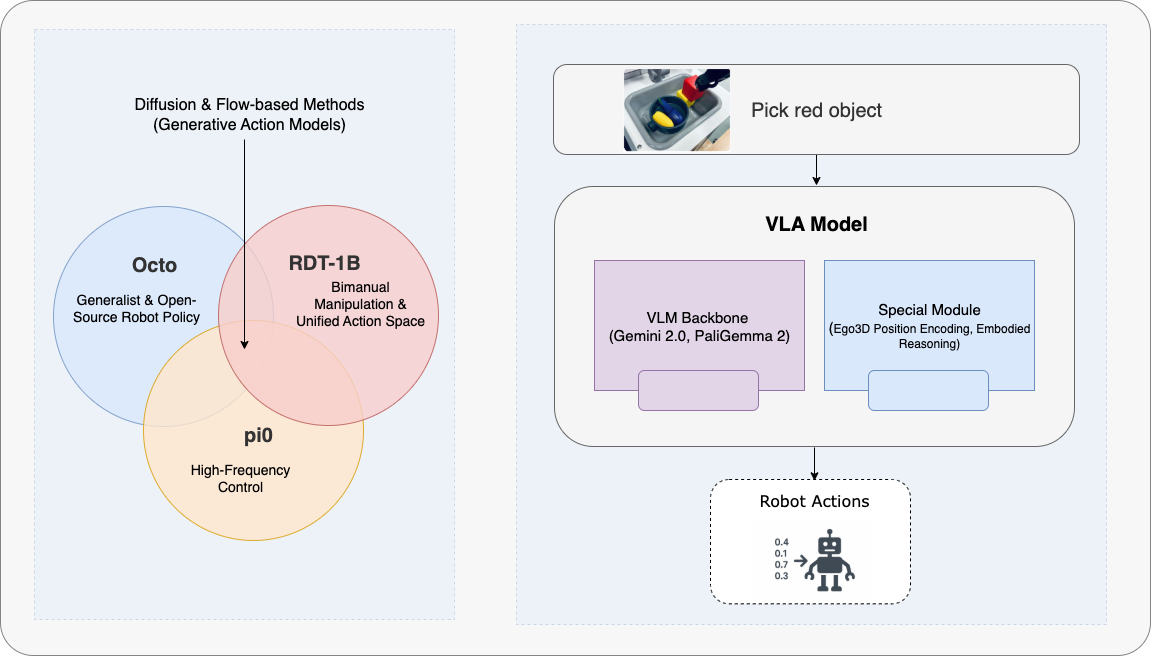} 
\caption{Left: Advanced Action Modeling - This group of models uses generative models for action representation, Right: Spatial and Embodied Reasoning - This group of works goes beyond basic visual inputs by incorporating a deeper understanding of 3D space and physical relationships (relevant works: SpatialVLA and Gemini Robotics)}
\label{fig:vla_tech}
\end{figure*}

Beyond discretization, a second strand of research embraces diffusion and flow-matching techniques to model the continuous distribution of robot actions. CogACT \cite{li2024cogact} exemplifies this separation of concerns: it uses a pretrained VLM for perception but delegates trajectory generation to a diffusion-based “cognition-to-action” block, capturing multimodal and temporally correlated dynamics. RDT-1B \cite{liu2024rdt} extends this to bimanual manipulation, using a diffusion transformer to predict long action horizons (64 steps) and showing nearly double the performance of models trained without large-scale pretraining. $\pi_0$ \cite{black2024pi_0} takes the flow-matching route (as shown in Figure \ref{fig:vla_tech}), parameterizing an ODE-based transformation on noisy action samples to generate smooth high-frequency control signals (up to 50 Hz). Collectively, these methods address the precision bottlenecks of autoregressive tokenization, enabling dexterity and responsiveness.

\begin{figure*}[htbp] % htbp stand for "here", "top", "bottom", "page"
\includegraphics[scale=0.35]{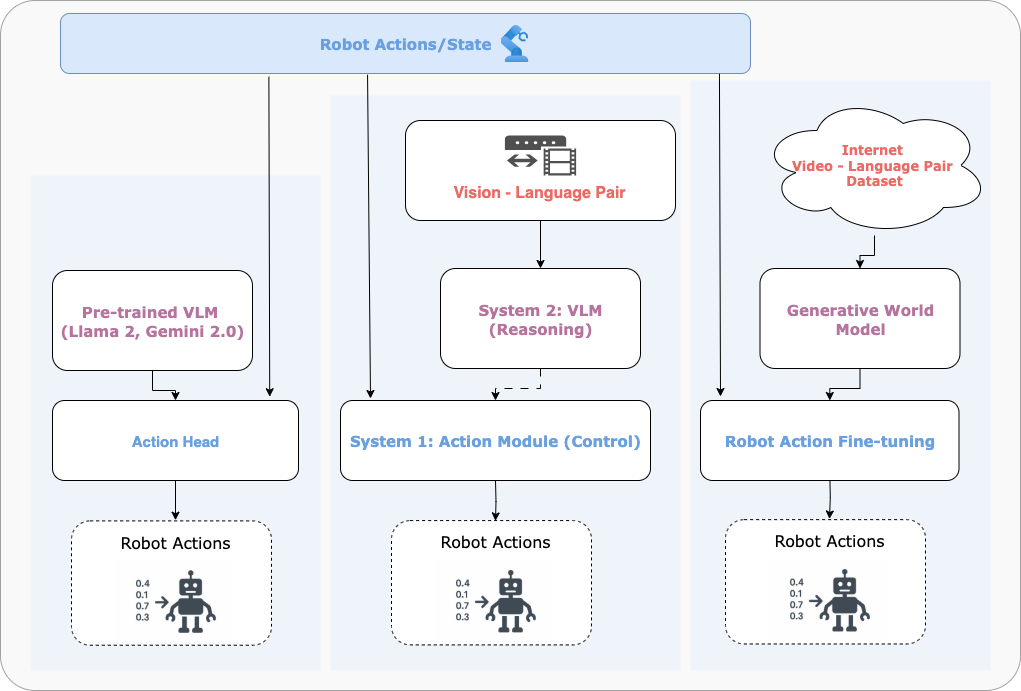} 
\caption{This categorizes the models into three groups based on their core design philosophy, showing a high-level view of how each model approaches the problem of VLA modeling. Left: directly adapting a pre-trained VLM for robotic control (relevant works: OpenVLA, Octo, Gemini Robotics, SpatialVLA), Middle: Separates the high-level reasoning from low-level action generation (relevant works: CogACT, GROOT N1, CoT-VLA), Right: Focuses on pre-training on massive datasets of non-robotic videos to learn the underlying dynamics of the world, a form of "embodied physics," before specializing for robot control. (relevant works: GR-2)}
\label{fig:vla_arch}
\end{figure*}

A complementary direction introduces cognitive and data-centric innovations. GR00T N1 \cite{bjorck2025gr00t} adopts a dual-system design (shown in Figure \ref{fig:vla_arch}: a pretrained VLM (System 2, \textasciitilde{1.34}B parameters) interprets vision-language input at low frequency (\textasciitilde{10} \textit{Hz}), while a diffusion-based action transformer (System 1, \textasciitilde{0.9}B parameters) outputs motor commands at high frequency (\textasciitilde{120} \textit{Hz}) using flow matching. Its “data pyramid” strategy, integrating internet video, synthetic physics simulations, and real-world robot data, provides robustness to the scarcity and heterogeneity of robot demonstrations. Similar cognition-inspired modularity underlies models like CogACT, which demonstrate substantial improvements over both OpenVLA and RT-2 on standard benchmarks.

In contrast to these works, the GR-2 \cite{cheang2024gr} model adopts a video-first approach. It is pre-trained on a massive volume of internet videos, specifically, 38 million video clips, to predict future frames autoregressively. This process is designed to enable the model to acquire a deep understanding of the world's dynamics, which is then transferred to downstream policy learning during a subsequent fine-tuning stage for action prediction. This methodology treats video prediction as a form of world modeling, providing a strong prior for understanding temporal and physical interactions before learning to control a robot.

Taken together, these advances illustrate a field maturing beyond the early era of discretized action tokens. The emerging consensus is that scalable perception (via VLM backbones and massive datasets) must be coupled with expressive, continuous action modeling (via spatial structures, diffusion, or flow-based architectures). This evolution is transforming VLAs from proof-of-concept systems into dexterous, generalist robot policies with the ability to reason, adapt, and act in complex real-world environments.

The rapid progression of VLA models showcases a significant shift from simple fine-tuning to sophisticated multi-component architectures that deeply integrate visual and linguistic knowledge. These large-scale learning resources are enabling robots to achieve unprecedented dexterity and robust generalization by mastering complex physical and semantic reasoning. Table \ref{tab:vla_models} shows the comparative analysis of the discussed VLA models.

\begin{table*}[h!]
\centering
\footnotesize
\begin{tabular}{p{2.0cm} p{3.1cm} p{3.6cm} p{4.5cm}}
\toprule
\textbf{Model} & \textbf{Core VLM/Pre-training} & \textbf{Action Represent. Method} & \textbf{Key Architectural Innovation} \\
\midrule

OpenVLA \cite{kim2024openvla} & Llama 2 + DINOv2 \& SigLIP & Discrete Tokenization & Open-source VLA, efficient fine-tuning (LoRA) \\
\midrule

Octo \cite{team2024octo} & From scratch (Transformer-first) & Diffusion-based Modeling & Flexible, compositional architecture, open-source \\
\midrule

CogACT \cite{li2024cogact} & Prismatic VLM (Llama 2 + DINOv2 \& SigLIP) & Diffusion Action Transformer & Decoupling of "cognition" (VLM) and "action" (DiT) \\
\midrule

RDT-1B \cite{liu2024rdt} & From scratch (Diffusion Transformer) & Diffusion-based Modeling & Specialized for bimanual control, Unified Action Space \\
\midrule

$\pi_0$ \cite{black2024pi_0} & PaliGemma 2 & Flow Matching & High-frequency control (50 \textit{Hz}), multi-expert architecture \\
\midrule

GROOT N1 \cite{bjorck2025gr00t} & Gemini 2.0 (distilled) & Diffusion Transformer & Dual-system architecture (System 2 + System 1), data pyramid \\
\midrule

SpatialVLA \cite{qu2025spatialvla} & PaliGemma 2 & Adaptive Action Grids & Ego3D Position Encoding, spatial-aware action tokenization \\
\midrule

GR-2 \cite{cheang2024gr} & From scratch on internet video & Conditional VAE & Video-generative pre-training, whole-body control (WBC) \\
\midrule

CoT-VLA \cite{zhao2025cot} & VILA-U (generative multimodal) & Hybrid: Causal + Full Attention & Visual chain-of-thought (CoT) reasoning with subgoal images \\
\midrule

Gemini Robotics \cite{team2025gemini} & Gemini 2.0 (VLM) & VLA on Gemini Robotics-ER & Dedicated embodied reasoning model (ER) for perception \\

\bottomrule
\end{tabular}
\caption{Comparative overview of state-of-the-art VLA models }
\label{tab:vla_models}
\end{table*}

\section{Approaches to Learning from Videos}
\label{sec:approaches}

\begin{figure*}[htbp] % htbp stand for "here", "top", "bottom", "page"
\includegraphics[scale=0.45]{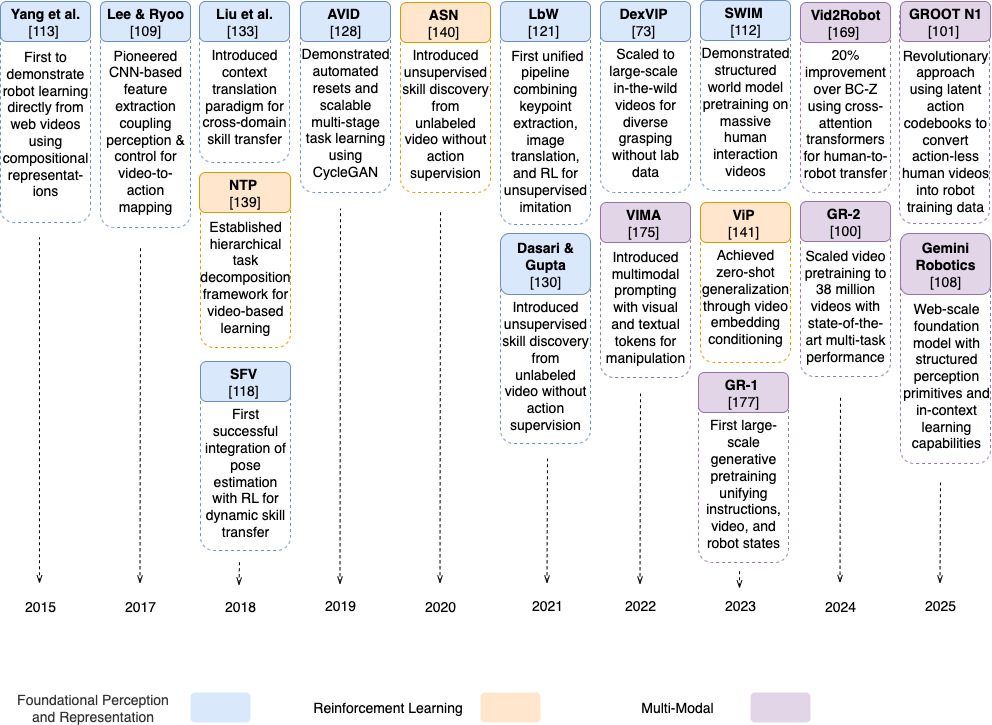} 
\caption{A comprehensive timeline organized chronologically by publication year, highlighting the key breakthroughs and milestones each work introduced to the field.}
\label{fig:timeline}
\end{figure*}

Researchers have proposed several approaches that adapt videos as the
data source for training robots for manipulation tasks. Some of these
approaches have borrowed many ideas from computer vision, while a few
have also incorporated ideas from language modeling. Substantial advancements have been documented within this domain; nevertheless, a more profound comprehension of the issue and additional exploration into novel learning methodologies, as well as fine-tuning existing ones, are needed to enhance the manipulation skills acquired by robots. The resulting discussions critically assess important literature in this field, highlighting prevailing challenges that impede the capacity of robots to acquire manipulation skills through
passive video observation.

The field of video-based robot manipulation encompasses a variety of
approaches, each leveraging different techniques to enable robots to
learn and perform tasks by observing video demonstrations. This
section categorizes these approaches into distinct but interrelated
groups, providing a coherent framework for understanding the landscape
of methods in this domain. The timeline of the most impactful works and breakthroughs presented in this review is illustrated in Figure \ref{fig:timeline}.

\subsection{Foundational Perception and Representation Methods}
\label{sec:foundational_methods}
Effective robot learning from videos relies on establishing perceptual and representational foundations that capture relevant task information and facilitate transfer across domains. On one hand, feature extraction methods focus on deriving meaningful visual or structural representations from raw video, using techniques such as CNN-based encoding or pose and keypoint detection. On the other hand, domain bridging approaches address the gap between human demonstration videos and robotic execution by translating visual and contextual information into robot-compatible representations. Together, these methods provide the essential perceptual groundwork for higher-level learning from videos.

\subsubsection{Feature Extraction Methods}
\label{sec:feature_extraction}
\begin{figure}[htbp] % htbp stand for "here", "top", "bottom", "page"
\includegraphics[scale=0.27]{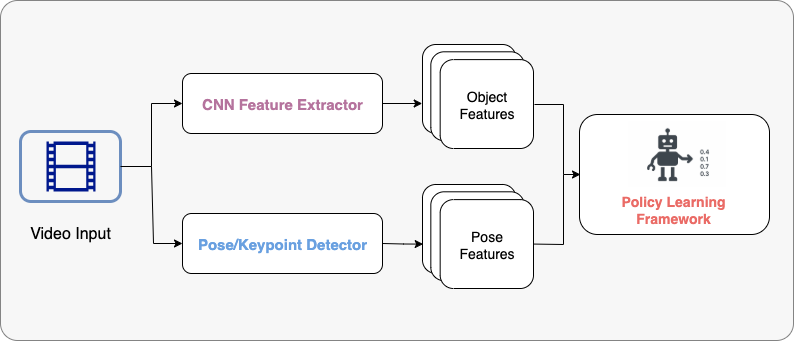} 
\caption{Feature extraction methods in video-based robot learning. CNN-based pipelines extract object features and masks, while pose/keypoint methods capture skeletal motion cues, both providing intermediate representations for policy learning.}
\label{fig:feature_extraction}
\end{figure}

A key step in learning from video is extracting task-relevant features that can guide robot policies. CNN-based approaches learn spatio-temporal representations directly from raw pixels for tasks such as object detection and hand-object interaction. In contrast, pose estimation and keypoint detection methods provide higher-level structural cues by explicitly modeling object geometry and motion dynamics. These two complementary strategies form the basis of video feature extraction for robot learning, as discussed below and shown in Figure \ref{fig:feature_extraction}.

\begin{itemize}
    \item \textbf{CNN-based Feature Extraction}:
    \label{sec:cnn_based}
    Early efforts in video-based robot learning primarily relied on Convolutional Neural Networks (CNNs) to extract visual features from demonstration videos, laying the foundation for mapping raw perception to robot action. A representative example is the work of \cite{lee2017learning}, which combined a CNN-based Single Shot MultiBox Detector (SSD) for hand-object interaction detection with a fully convolutional network (FCN) to predict future hand positions. By coupling perception and control, the system directly translated visual cues into motor commands, enabling adaptive robot behavior.

    Subsequent approaches expanded this pipeline with richer perception modules to better capture the complexity of human demonstrations. For instance, \cite{zhang2019learning} augmented CNN-based object detection with OpenPose-based hand localization and greedy video segmentation, allowing the robot to infer collaborative actions and object transfers from relative spatial configurations. Similarly, \cite{zhang2019object} integrated two-stream CNNs with Mask R-CNN to construct a video parser, which, in combination with a grammar-based execution module, translated visual observations into structured manipulation commands. These approaches illustrate a shift from simple object detection toward frameworks that combine visual parsing with higher-level action reasoning.
    
    More recent works have pushed CNN-based pipelines toward generalization and scalability. The SWIM framework \cite{Mendonca2023StructuredWM}, for example, pretrains on large-scale human interaction videos to learn structured world models of hand-object manipulation. With minimal finetuning on robot data, these representations support goal-conditioned planning, demonstrating the utility of pretraining as a bridge from human demonstrations to robot execution. Previous efforts, such as \cite{yang2015robot}, emphasized compositional representations by pairing CNN-based object detection with grammar-based parsers, highlighting the move toward language-like abstractions of manipulation.
    
    Addressing domain shift, \cite{schmeckpeper2020learning} introduced latent variable models trained on both labeled and unlabeled videos to disentangle shared and domain-specific factors of action. More recently, \cite{ko2023learning} took this further by eliminating the need for action labels, synthesizing robot-action videos from demonstrations and learning policies directly from raw RGB inputs. These advances mark a trajectory from frame-level feature extraction to structured, generalizable, and annotation-free models of robot manipulation, steadily expanding the robustness and applicability of video-based learning.
    \\
    \item \textbf{Pose and Keypoint Detection}:
    \label{sec:pose_based}
    While CNN-based pipelines emphasized holistic visual features, a parallel line of work has focused on pose and keypoint detection to capture the fine-grained structure of human motion. Instead of treating video frames as undifferentiated inputs, these methods identify joints, fingertips, or object centers and assemble them into coherent spatial arrangements, providing robots with precise cues about the intent and feasibility of demonstrated actions.

    Several studies leveraged pose estimation to bridge the gap between human demonstrations and robot control directly. For example, \cite{bahl2022human} extracted detailed hand poses using the 100DOH model and mapped them into robot coordinates with depth sensing, creating strong priors for policy learning. Similarly, \cite{sivakumar2022robotic} proposed a two-stage pipeline that first reconstructs 3D human poses and then retargets them to robot kinematics, ensuring that trajectories remain both functional and feasible. These approaches emphasize embodiment alignment, translating human movements into robot-compatible actions.
    
    Pose-based reasoning has also proven valuable for dynamic skill transfer. The work of \cite{peng2018sfv} reconstructed actor trajectories from monocular videos and trained reinforcement learning controllers to imitate these motions in simulation, demonstrating how weakly supervised pose estimation can bootstrap skill acquisition. Domain-specific applications, such as \cite{deng2022human}, have fused Keypoint-RCNN with imitation learning to teach assembly tasks, where reconstructed hand trajectories guided a UR3 robot through precise manipulation.
    
    Scaling beyond curated datasets, methods like DexVIP \cite{mandikal2022dexvip} exploit large-scale "in-the-wild" human-object interaction videos to learn diverse grasping strategies. By combining human hand pose priors with reinforcement learning, DexVIP achieved generalizable grasping without costly lab-collected data. Meanwhile, lighter-weight approaches combine keypoints with domain adaptation: \cite{sun2022learning} used EfficientNet-based keypoint detection with CycleGAN translation to align human and robot domains, enabling learning from raw demonstrations. The Learning by Watching (LbW) framework \cite{xiong2021learning} further unified keypoint extraction, image translation, and reinforcement learning into a pipeline that enables unsupervised imitation from human videos without expert annotation.
    
    Together, these works illustrate an evolution from basic pose/keypoint detection toward pipelines that integrate retargeting, domain adaptation, and large-scale pretraining. By capturing structured motion cues, pose-based methods complement CNN-based perception, offering robots interpretable, transferable representations of human actions for imitation and skill learning.
\end{itemize}

\begin{table*}[h!]
\centering
\footnotesize
\begin{tabular}{p{0.8cm} p{2.5cm} p{2.5cm} p{2.4cm} p{2.4cm} p{2.5cm}}
\toprule
\textbf{Method} & \textbf{Task Performance} & \textbf{Sample Efficiency} & \textbf{Compute Cost} & \textbf{Embodi. Handling} & \textbf{Pros (+) / Cons (-)} \\
\midrule

\cite{lee2017learning} & Good; real-time collaborative tasks & Moderate; unlabeled videos, needs hand annotations. & Moderate (CNN-based networks) & Direct transfer by visual regression & (+) Unlabeled human videos; (-) Needs initial annotations \\
\midrule

\cite{peng2018sfv} & Robust, generalizable dynamic skills (e.g., acrobatics) & Moderate-high; leverages abundant video data & High (deep RL, pose estimation, and simulation) & Pose estimation/motion reconstruction. & (+) Rich video data, dynamic skills; (-) High compute and reconstruction complexity \\
\midrule

\cite{schmeckpeper2020learning} & Effective tool-use learned purely from observation & Moderate; leverages both observation and interaction data & Moderate-high (latent models and inference) & Learns domain-specific priors & (+) Passive human obs., generalizable; (-) Needs careful latent modeling \\
\midrule

\cite{xiong2021learning} & Effective simulation manipulation tasks & Moderate; relies on unsupervised translation and detection & Moderate (images translation, keypoints detection, and RL) & Unsupervised human-to-robot translation & (+) Structured semantics, unsupervised; (-) Simulation only \\
\midrule

\cite{mandikal2022dexvip} & Effective dexterous grasping in simulation & High; leverages video priors & Moderate-high (RL, pose extraction) & Hand-pose priors from video & (+) Pose priors, good generalization; (-) Pose extraction limits \\
\midrule

\cite{sivakumar2022robotic} & Robust teleoperation \& dexterous control in real-time & Low; uses large unlabeled videos & Moderate (pose estimation, neural retargeting) & 3D pose-based retargeting & (+) Real-time, low data; (-) Needs reliable pose estimation \\
\midrule

\cite{bahl2022human} & Good; generalizes in many manipulation tasks & High; learns from single human demonstration & Moderate (sampling-based optimization, alignment loss) & Human priors + video alignment & (+) One-shot learning; (-) Sensitive to priors \\
\midrule

\cite{Mendonca2023StructuredWM} & Robust for various manipulation tasks & High; few real-world trajectories needed & Moderate-high (world model training and finetuning) & Affordance learning & (+) Few-shot, robust; (-) High model training cost \\
\midrule

\cite{ko2023learning} & Effective performance across manipulation/navigation & High; no action labels needed & Moderate (video synthesis, flow prediction, optimization) & Dense correspondences for action & (+) Inference without action labels; (-) Relies on flow accuracy \\
\midrule

\cite{yang2015robot} & High accuracy on action parsing from unconstrained videos & Moderate; uses large video sets & Moderate (CNN, grammar parsing) & Perception module based & (+) Handles unconstrained data; (-) Needs reliable perception \\
\midrule

\cite{deng2022human} & Effective in collaborative assembly tasks & Moderate; needs multiple demonstrations & Moderate-high (pose estimation nets, trajectory optimization) & Pose estimation + video retargeting & (+) Accurate pose, task alignment; (-) Video/camera quality sensitive \\
\midrule

\cite{zhang2019object} & Good in multi-object manipulation & High; uses attribute-guided demos & Moderate (CNN, attribute inference) & Attribute-specific retargeting & (+) High object-specific accuracy; (-) Needs attribute extraction \\
\midrule

\cite{sun2022learning} & Effective in simple manipulation tasks & High; 20-30 demos sufficient & Moderate (keypoints extraction, CycleGAN, BC, SQIL) & CycleGAN-based translation & (+) Efficient, robust; (-) Limited task complexity \\
\midrule

\cite{zhang2019learning} & Good in collaborative parsing from unconstrained video & Moderate; uses public unlabeled videos & Moderate (YOLO, OpenPose, grammar) & Grammar/symbolic parsing & (+) Generalizable parsing; (-) Action recog. accuracy limits \\
\bottomrule
\end{tabular}
\caption{Comparison of feature extraction-based approaches for learning manipulation skills from human video}
\label{tab:feature_extract_video_methods}
\end{table*}

Table \ref{tab:feature_extract_video_methods} shows that feature extraction methods are data-efficient and flexible, well-suited for leveraging large unlabeled video datasets and producing interpretable representations. They work especially well with reinforcement or imitation learning but face challenges with embodiment transfer, scaling to long-horizon reasoning, and domain gaps. Thus, they are valuable for action parsing, object-centric manipulation, and collaboration, though less effective for abstract policy reasoning or seamless transfer.

\subsubsection{Domain Bridging via Translation}
\label{sec:image_context_translation}
When robots learn manipulation skills from human demonstration videos, a significant challenge arises due to the domain shift, primarily stemming from differences in embodiment, visual appearance, and context between humans and robots. Bridging this gap requires intermediate representations and translation techniques that enable robust transfer of skills across domains. Image and context translation methods have emerged as effective solutions, focusing on transforming visual and contextual information to enhance skill transfer, generalization, and adaptability for robotic manipulation.

\begin{itemize}
    \item \textbf{Image Translation}:
    \label{sec:image_translation}
    A major challenge in video-based robot learning is the domain gap between human demonstrations and robot execution. Image translation methods address this by transforming visual data across domains, for instance, converting human-centric demonstrations into robot-perspective images or adapting synthetic renderings into realistic ones. By aligning visual appearances, these methods create a bridge that allows robots to directly interpret and imitate human actions. Classic GAN-based approaches \cite{isola2017image,zhu2017toward,liu2017unsupervised} laid the foundation, enhancing the realism of training data and improving robots’ ability to generalize manipulation skills \cite{erdei2024image}.

    Within robotics, translation has proven especially powerful for mapping fine-grained object and hand interactions into robot embodiments. For example, \cite{sharma2019third} introduced a two-module pipeline where a conditional GAN (and domain-invariant networks \cite{schmeckpeper2020learning,finn2017model,sermanet2018time,chen2021learning,yu2018one}) predicted visual sub-goals from demonstration videos, while a low-level controller generated corresponding actions. This decoupling of perception and control allowed skill transfer even with unaligned datasets. Similarly, Automated Visual Instruction-following with Demonstrations (AVID) \cite{smith2019avid} employed CycleGAN \cite{zhu2017unpaired} to translate human videos directly into robot images at the pixel level, producing instruction images that served as reward signals for reinforcement learning. These works highlighted how direct appearance-level alignment could eliminate the need for manual correspondence between human and robot demonstrations.
    
    Recent advances have built on these foundations with more expressive architectures. Transformer-based systems, such as the one-shot imitation framework in \cite{dasari2021transformers}, leveraged self-attention modules to perform unsupervised image-to-image translation, combining goal-conditioned behavioral cloning with deep RL to robustly track and imitate behaviors. Beyond pixel-level mappings, meta-imitation learning methods like A-CycleGAN \cite{li2022metaimitation} introduced bi-directional translation between human and robot domains. Coupled with self-adaptive meta-learning, these systems generate imagined robot data to support rapid adaptation with minimal demonstrations, marking a shift toward scalable and flexible cross-domain imitation.
    \\
    \item \textbf{Context Translation}:
    \label{sec:context_translation}
    Whereas image translation focuses on aligning visual appearance, context translation tackles the broader challenge of transferring skills across tasks, environments, and viewpoints. This enables robots to adapt behaviors learned in one setting to novel situations with different backgrounds, object positions, or camera perspectives \cite{liu2023ceil}.

    One of the earliest examples is \cite{liu2018imitation}, which trained a context translation model on paired demonstrations from diverse scenarios. By learning to predict how the same skill looks across contexts, the system enabled a robot to reproduce behaviors in new environments using reinforcement learning. Building on this, \cite{yang2020cross} developed a context-agnostic task representation paired with a multi-modal inverse dynamics model. By fusing RGB and depth data (and compensating when depth was missing), their system achieved robust action prediction across diverse viewpoints and object configurations.
    
    Integration with unsupervised video translation further enhanced scalability. The Learning by Watching (LbW) framework \cite{xiong2021learning} combined MUNIT-based \cite{huang2018multimodal} video translation with unsupervised keypoint detection, mapping human demonstrations into robot domains without explicit task supervision. Structured keypoint representations extracted from translated videos served as inputs for reinforcement learning, enabling robots to imitate behaviors under varying contextual constraints. Crucially, both training data for humans and robots were collected via random demonstrations rather than expert labels, lowering the barrier for large-scale data collection.
\end{itemize}

Image and context translation offer complementary strategies to bridge the human-robot domain gap. Image translation aligns the robot’s visual perception with demonstrations, while context translation adapts actions across environments, viewpoints, and tasks. Progress has evolved from early GAN-based pixel translation to transformer and meta-learning frameworks, moving toward scalable, annotation-efficient, and robust pipelines for unstructured settings.

\begin{table*}[h!]
\centering
\footnotesize
\begin{tabular}{p{0.8cm} p{2.5cm} p{2.5cm} p{2.4cm} p{2.4cm} p{2.5cm}}
\toprule
\textbf{Method} & \textbf{Task Performance} & \textbf{Sample Efficiency} & \textbf{Compute Cost} & \textbf{Embodi. Handling} & \textbf{Pros (+) / Cons (-)} \\
\midrule

\cite{liu2018imitation} & Good; high sim task success; outperforms GAIL/TPIL & Moderate; needs multiple human video demos from varied contexts and 100k+ samples during RL & Moderate; translation model + visual encoder + RL (TRPO or GPS) & Weak; assumes human and robot use same tools and similar viewpoint to reduce domain gap & (+) Learns from raw video; (-) Requires morphology/demo alignment \\
\midrule

\cite{sharma2019third} & High; 60-75\% real robot success; outperforms end-to-end and DAML baselines & High; reusable low-level controller; fewer task-specific samples needed & Moderate; U-Net + ResNet-based goal generator + inverse model; trained separately & Moderate; uses GAN-based sub-goal transfer & (+) Hierarchical; modular/sample-efficient; (-) Goal generator task-specific \\
\midrule

\cite{smith2019avid} & High; 80-100\% multi-stage task success & Very high; learns full tasks with \textasciitilde{20} mins of human video and 180 mins of robot practice & High; CycleGAN + structured latent model + MPC + classifier-based reward & Strong; translates entire human demo to robot via CycleGAN without paired data & (+) Automated resets, scalable; (-) CycleGAN requires good translation data, per-task model \\
\midrule

\cite{dasari2021transformers} & High; 88.8\% pick-place success & High; 3x less data needed & High; Transformer with inverse dynamics and keypoint loss & Strong; self-supervised sim-to-real & (+) Excellent transfer, modular; (-) High data/compute \\
\midrule

\cite{xiong2021learning} & Strong; comparable or superior to AVID/Classifier-based methods & High; learns from single human demo per task & Moderate; CycleGAN + keypoint extraction + RL & Strong; keypoints structure for transfer & (+) Avoids artifacts, efficient; (-) Needs robust translation model \\
\midrule

\cite{li2022metaimitation} & Strong; matches DAML without robot demos & High; only human video needed during training & High; A-CycleGAN + meta-learning & Strong; A-CycleGAN handles shifts & (+) No robot demos; (-) Relies on good latent/action inference \\
\midrule

\cite{yang2020cross} & Good; beats baselines on stacking in different contexts & Moderate; uses paired data and depth prediction & Moderate-high; 4-model pipeline with depth estimation & Moderate; via context translation and multimodal input & (+) Strong cross-context performance with RGB-D; (-) Needs depth prediction model and alignment \\
\bottomrule
\end{tabular}
\caption{Comparison of image and context translation-based approaches for learning manipulation skills from human video}
\label{tab:image_context_video_methods}
\end{table*}

Table \ref{tab:image_context_video_methods} illustrates how these methods address visual and semantic gaps, enabling sim-to-real and cross-domain transfer from raw or unpaired video data. However, challenges remain in precise action alignment, artifact reduction, and managing embodiment mismatches. Despite these hurdles, context translation remains critical for advancing generalization and real-world adaptability.

\subsection{Reinforcement Learning (RL) Approaches}
\label{sec:rl_based}
\begin{figure*}[htbp] % htbp stand for "here", "top", "bottom", "page"
\includegraphics[scale=0.40]{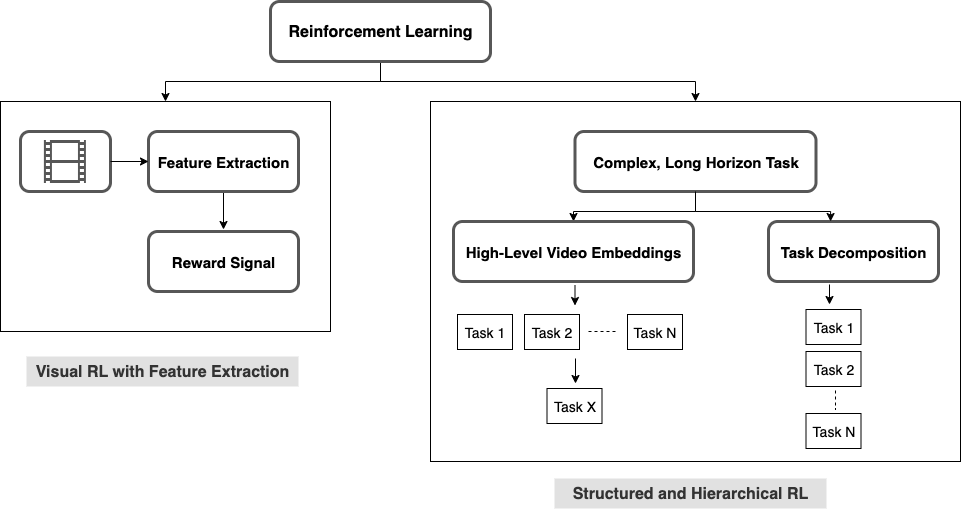} 
\caption{Reinforcement learning paradigms in video-based robot learning - Left: visual RL with feature extraction, grounding policies in parsed video features; Right: structured and hierarchical RL, learning high-level video embeddings for multiple tasks, and decomposing long-horizon tasks into subtasks and primitive skills.}
\label{fig:rl}
\end{figure*}

Reinforcement Learning (RL) provides a powerful paradigm for enabling robots to acquire manipulation skills through interaction, optimizing behavior by trial and error guided by reward signals. In the context of learning from video, recent research has explored how RL can be combined with rich perceptual inputs extracted from demonstrations. This integration allows robots to benefit both from their exploratory learning and from structured guidance provided by human expertise in video data. A depiction of the RL-based subcategories is shown in Figure \ref{fig:rl}.

\subsubsection{Visual RL with Feature Extraction}
\label{sec:visual_rl_feat_extract}

Early approaches sought to ground RL in visual representations derived from video, leveraging feature extraction to transform demonstrations into useful training signals. For example, the video parsing framework of \cite{petrik2021learning} combined Mask R-CNN with a dedicated hand-object detector to build coarse 3D scene representations from human demonstrations. These representations were aligned across trajectories and used to generate dense reward signals, guiding RL policies toward precise motor execution. Similarly, \cite{zorina2021learning} emphasized extracting tool motion from instructional videos, aligning simulated environments with human demonstrations, and employing trajectory optimization to bridge from visual guidance to executable robot policies. In both cases, parsing video into trajectories provided structured signals that made RL training more efficient.

A persistent challenge, however, is that raw video demonstrations typically lack explicit action or reward labels, and domain shift can make learned policies brittle. To address this, \cite{schmeckpeper2021reinforcement} introduced a hybrid approach that combined offline observational data with online interaction. Their system maintained dual replay buffers, one for action-free video observations and another for action-conditioned robot experience, and learned inverse models over compressed, domain-invariant features. This design allowed robots to infer missing actions and transfer knowledge across domains, advancing RL toward robustness in heterogeneous video settings.

\subsubsection{Structured and Hierarchical RL}
\label{sec:structured_and_hierarchical_rl}

As research progressed, focus shifted from single-task policies toward generalizable and scalable frameworks. Neural Task Programming (NTP) \cite{xu2018neural} exemplified this trend by decomposing tasks into hierarchical compositions of primitive actions, parameterized by trajectories or video demonstrations. Through meta-learning strategies, NTP enabled robots to adapt policies quickly across diverse manipulation tasks. Likewise, Adversarial Skill Networks (ASN) \cite{mees2020adversarial} learned a task-agnostic skill embedding space from unlabeled, multi-view observations. By leveraging adversarial and metric learning objectives, ASN eliminated the need for explicit action supervision, highlighting how representation learning can facilitate transferable skills.

More recently, Video-conditioned Policy Learning (ViP) \cite{chane2023learning} has advanced the integration of RL with large-scale video data. ViP conditions policies directly on video embeddings of demonstrations, using a supervised contrastive encoder trained on human activity datasets like Something-Something-v2 \cite{goyal2017something}. At inference, ViP retrieves task embeddings via nearest-neighbor search and conditions policy learning on them, enabling multi-task and zero-shot generalization. By directly grounding RL in high-level video representations, ViP points toward scalable solutions where robots can learn broad repertoires of behaviors from human video libraries without paired training data.

\begin{table*}[h!]
\centering
\footnotesize
\begin{tabular}{p{0.8cm} p{2.5cm} p{2.5cm} p{2.4cm} p{2.4cm} p{2.5cm}}
\toprule
\textbf{Method} & \textbf{Task Performance} & \textbf{Sample Efficiency} & \textbf{Compute Cost} & \textbf{Embodi. Handling} & \textbf{Pros (+) / Cons (-)} \\
\midrule

\cite{xu2018neural} & High; excels in hierarchical tasks (stacking, sorting) & High; few demos needed & Moderate (due to hierarchical decomposition and recursive calls) & Good; task decomposition, program induction & (+) Hierarchical generalization, modular, few-shot; (-) Sensitive to low-level API/collision \\
\midrule

\cite{mees2020adversarial} & High; learns complex tasks from video & Moderate; uses unlabeled video data & Moderate (metric and adversarial learning) & Good; adversarial skill-transfer embeddings & (+) Transferable embeddings, unsupervised skill disc.; (-) Needs careful metric learning \\
\midrule

\cite{schmeckpeper2021reinforcement} & High; strong on vision-based tasks & High; leverages human videos, less robot data & Moderate (inverse model training, adversarial domain confusion) & Excellent; domain-invariant embedding & (+) Good domain generalization, sim-to-real; (-) Dependent on inverse model accuracy \\
\midrule

\cite{petrik2021learning} & Good; effective for object manipulation tasks & Moderate; few demo videos needed & Moderate (differentiable rendering, RL training) & Excellent; 3D state estimation transfer & (+) Robust 3D generalization; (-) Approximation errors for complex scenes \\
\midrule

\cite{zorina2021learning} & High; effective for tool manipulation (spade, hammer, scythe), 100\% success rate & High; requires only single video demonstration & Moderate (trajectory optimization + PPO, alignment sampling up to 20k iterations) & Excellent; morphology-agnostic via tool-centric approach & (+) Tool-focused transfer, works across robot morphologies, real robot validation; (-) Requires sparse reward environment, limited to stick-like tools, needs good video visibility \\
\midrule

\cite{chane2023learning} & Excellent; zero-shot manipulation from human videos & Highly efficient; leverages large human datasets & Moderate (pretrained video embedding, efficient inference) & Excellent; pretrained action embeddings & (+) Zero-shot generalization, efficient inference; (-) Relies on pretrained embeddings \\
\bottomrule
\end{tabular}
\caption{Comparison of reinforcement learning-based approaches for learning manipulation skills from human video}
\label{tab:rl_video_methods}
\end{table*}

Table \ref{tab:rl_video_methods} highlights the strength of reinforcement learning in tackling complex, long-horizon tasks and achieving robust sim-to-real transfer, particularly in hierarchical and multi-task manipulation. However, these gains come with high computational and data costs, along with sensitivity to reward design and training stability, limiting RL’s practicality despite its adaptability.

\subsection{Imitation Learning (IL) Approaches}
\label{sec:il_based}
\begin{figure*}[htbp] % htbp stand for "here", "top", "bottom", "page"
\includegraphics[scale=0.52]{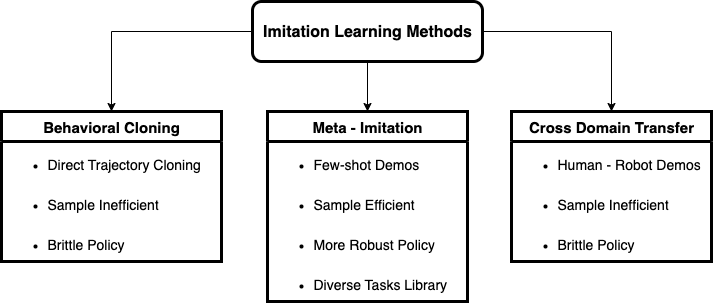} 
\caption{Taxonomy of imitation learning approaches - Left: behavioral cloning from raw videos, (Middle): meta-imitation from few-shot demonstrations, and Right: cross-domain transfer from human-to-robot videos.}
\label{fig:il}
\end{figure*}

Imitation Learning (IL) is one of the most prominent strategies for training robot manipulation policies. In contrast to reinforcement learning, which depends on trial-and-error exploration, IL enables robots to acquire skills directly from human demonstrations. This reduces the need for carefully engineered reward functions and often yields more sample-efficient, generalizable policies, particularly valuable in data-limited or real-world settings. Figure \ref{fig:il} shows the different taxonomy of IL based approaches, and they are discussed in details below.

\subsubsection{Behavioral Cloning (BC) and Variants}
\label{sec:bc_and_variants}
A large body of IL research builds on behavioral cloning (BC), where visual inputs from demonstration videos are mapped directly to robot actions. Modern approaches typically employ CNN-based encoders for feature extraction, combined with deep policy networks for action prediction.

Recent extensions of BC have introduced more flexible task representations. For example, BC-Z \cite{jang2022bc} supports both video- and language-specified tasks, enabling zero-shot generalization. Its architecture combines a ResNet18 \cite{he2016deep} backbone with FiLM layers \cite{perez2018film} for task conditioning, and incorporates human-in-the-loop corrections via teleoperation.

Other works have focused on data efficiency. TecNets \cite{bonardi2020learning, james2018task} encode demonstrations into compact embeddings that condition policies for rapid adaptation, while Multiple Interactions Made Easy (MIME) \cite{sharma2018multiple} scales imitation through a large demonstration dataset, pairing VGG-based \cite{simonyan2014very} visual encoders with LSTM-based \cite{hochreiter1997long} trajectory prediction.

Beyond direct cloning, IL can also be framed as an inverse reinforcement learning (IRL) problem, where the goal is to infer a cost function from demonstrations. A key direction here is Imitation from Observation (IfO), which avoids reliance on expert action labels. Generative Adversarial Imitation from Observation (GAIfO) \cite{torabi2018generative} exemplifies this approach by recovering expert-like policies from observed state transitions alone, providing a more scalable alternative to traditional IRL.

\subsubsection{Meta-Imitation and Few-Shot IL}
\label{sec:meta_il}
To improve generalization, recent research integrates IL with meta-learning and few-shot learning. The central idea is to train on diverse tasks so that robots can quickly adapt to novel ones from only a handful of demonstrations \cite{finn2017model, song2023comprehensive}.

Zero-shot imitation has been explored by \cite{pathak2018zero}, who combined exploration-driven policies with forward-consistency losses to enable imitation without labeled demonstrations. Similarly, MOSAIC \cite{mandi2022towards} employs multi-task architectures with self-attention and temporal contrastive modules, enabling robust representation learning and improved task disambiguation. Meta-learning methods such as Model-Agnostic Meta-Learning (MAML) have also been adapted for IL. \cite{yu2018one} trained policies across varied prior tasks so that only a single new demonstration is required for adaptation at inference. This one-shot generalization illustrates the promise of meta-imitation in tackling domain shift and data scarcity.

\subsubsection{Cross-Domain and Human-to-Robot Transfer}
\label{sec:hr_transfer_il}
A parallel line of work addresses cross-domain transfer, particularly from human video demonstrations to robot execution. The work done in \cite{goo2019one} demonstrated multi-step task learning by localizing human-demonstrated actions within supplemental videos, combining BC, IRL, and RL for accelerated refinement. The authors in \cite{kim2023giving} proposed a more direct strategy, applying BC on raw human videos without explicit domain adaptation. By leveraging end-to-end training with Adam optimization \cite{kingma2014adam} and occlusion-based perspectives (e.g., eye-in-hand views), their method mitigates domain shift and enables direct policy transfer. Other approaches incorporate auxiliary supervision signals. The Watch, try, learn (WTL) framework \cite{zhou2019watch} integrates visual meta-learning with binary success/failure feedback, enabling policy refinement in sparse-reward settings. Graph-based representations such as Visual Entity Graphs (VEGs) \cite{sieb2020graph} further enhance transfer by explicitly modeling spatial and temporal relations between objects and actions, enabling single-demonstration learning in real robotic systems.

\begin{table*}[h!]
\centering
\footnotesize
\begin{tabular}{p{0.8cm} p{2.5cm} p{2.5cm} p{2.4cm} p{2.4cm} p{2.5cm}}
\toprule
\textbf{Method} & \textbf{Task Performance} & \textbf{Sample Efficiency} & \textbf{Compute Cost} & \textbf{Embodi. Handling} & \textbf{Pros (+) / Cons (-)} \\
\midrule

\cite{torabi2018generative} & High, matches IL from state-only, no actions needed & Low; needs large robot data; not suitable for human demonstration or low-data settings & Moderate (model-free RL training with discriminators) & None; needs shared state/action space & (+) No action labels, partial demos; (-) No human-to-robot transfer, no raw vision use \\
\midrule

\cite{sharma2018multiple} & Not applicable; introduces a large-scale multi-task dataset & High; supports low-data learning algorithms & N/A (dataset paper) & Human/robot paired demos & (+) Largest video-trajectory dataset; (-) Limited real-world variety, kinesthetic only \\
\midrule

\cite{sieb2020graph} & Good; achieves robust performance using graph-structured perception from a single human video demonstration & High; one-shot imitation, no robot data required at test time & High (graph generation, matching over time, entity detectors) & Strong; dynamic graph aligns hand/object keypoints & (+) One-shot, no instrumentation; (-) Needs reliable keypoint detection, compute heavy \\
\midrule

\cite{bonardi2020learning} & High; effective at one-shot imitation in simulation using image-based embeddings & Moderate; needs diverse pairs for training; efficient at test time due to embedding reuse & Moderate (encoder + BC + embedding loss) & Partial; generalization by visual embedding & (+) General/one-shot; (-) Sensitive to domain gap, real-world untested \\
\midrule

\cite{mandi2022towards} & High, strong across 7 tasks, 61 using a unified transformer-based policy & Moderate; shared transformer, task data needed & High (transformer, temporal loss) & Partial; generalizes robot arms, not human-to-robot & (+) Unified policy, generalizes; (-) Complex pipeline, demo diversity required \\
\midrule

\cite{jang2022bc} & Good; 44\% zero-shot success on 24 unseen real-world tasks using video/language input & Moderate; requires thousands of demos, few/zero-shot efficient after pretraining & Moderate (ResNet encoders + FiLM conditioning + BC) & Strong; video/language embedding bridges modalities & (+) Flexible input, strong generalization; (-) Needs quality pretrained embeddings \\
\midrule

\cite{zhou2019watch} & Moderate, learns reward from online human video for RL & Low; video + environment interaction for reward inference and training & High (reward learning, video encoding, RL) & Weak; learning sensitive to visual mismatch & (+) Autonomous reward from video; (-) Artifacts/noise hurt policy \\
\midrule

\cite{yu2018one} & High; strong one-shot performance on sim and real-world robotic tasks using gradient-based adaptation & High; only one test-time demonstration required after meta-training on human+robot data & High (MAML optimization, visual encoder, adaptation at inference) & Moderate; domain-adaptive feature aligns human-robot & (+) Fast adaptation, flexible input; (-) Needs meta-training, paired modalities \\
\midrule

\cite{kim2023giving} & High; 58\% improvement in real-world manipulation & High; avoids expert robot demos by using inverse model trained on play data to label human videos & Moderate (inverse, BC, masking) & Strong; masking + camera perspective closes gap & (+) Robust/scalable, no robot demo; (-) Masking can omit context \\
\midrule

\cite{pathak2018zero} & High; real and simulation tasks via goal images, no expert actions & High; no expert demos, needs unsupervised pretraining & Moderate-high (exploration, forward consistency) & Moderate; learns goal-conditioned skills during exploration & (+) Unsupervised, strong generalization; (-) Relies on exploration/visual similarity \\
\midrule

\cite{goo2019one} & Moderate-High; achieves multi-step task execution from one segmented demo + auxiliary videos & High; few-shot learning via video snippets/meta-learned localization & Moderate; relies on few-shot video classification (MAML/Reptile) and PPO for policy training & Moderate; generalizes across unseen colors and robot arms, but not evaluated on human-to-robot transfer & (+) Multi-step one-shot, robust reward inference; (-) No real-robot transfer, sim only \\
\bottomrule
\end{tabular}
\caption{Comparison of imitation learning-based approaches for learning manipulation skills from human video}
\label{tab:il_video_methods}
\end{table*}

Imitation learning approaches, as summarized in Table \ref{tab:il_video_methods}, stand out for their sample efficiency and direct use of expert demonstrations, making them ideal for scenarios where rapid skill acquisition is critical. The comparative analysis shows that advances in meta and one-shot learning have further enhanced their ability to generalize from limited data. However, the table also highlights persistent issues with embodiment alignment and generalization, especially across domains or when transferring from human to robot. Many methods remain primarily evaluated in simulation. In practice, imitation learning excels for tasks with readily available demonstration data but may require augmentation or hybridization for broader applicability.

\subsection{Hybrid Approaches}
\label{sec:hybrid_method}

Hybrid models have emerged as a powerful paradigm in robot learning, combining RL, IL, and complementary techniques to overcome the limitations of relying on a single method. Early work demonstrated the effectiveness of augmenting IL with RL-based fine-tuning, such as attention-driven imitation guiding RL policies \cite{ramachandruni2020attentive} and pose-driven motion imitation refined via RL \cite{peng2018sfv}. RL-based residual corrections to demonstration trajectories \cite{garcia2020physics} has also stood out as an approach to adapt video demonstrations to robot trajectories. Other efforts leverage cross-domain priors \cite{pertsch2022cross}, unaligned video demonstrations \cite{aytar2018playing}, or adversarial third-person imitation \cite{stadie2017third} to improve generalization and robustness. More recent approaches combine IL, RL, and control-theoretic methods such as model predictive control (MPC) for more efficient skill transfer \cite{huo2023efficient}. By integrating multiple learning strategies, these approaches enhance data efficiency, adaptability, and robustness, particularly in complex manipulation tasks. 

\subsubsection{Hybrid RL and IL}
\label{sec:hybrid_rl_il}
A central motivation for hybrid RL-IL approaches is to combine the sample efficiency of imitation with the exploration and optimization strengths of RL. Attentive Task-Net \cite{ramachandruni2020attentive} exemplifies this direction by integrating a self-supervised attention network for viewpoint-invariant imitation with an RL policy optimized using Deep Deterministic Policy Gradient (DDPG) \cite{lillicrap2015continuous}. A CNN-based embedding network learns task-relevant visual features, which are refined by spatial attention and used to guide RL agents, producing policies that balance efficient imitation with long-horizon optimization. Similarly, Skills From Videos  (SFV) \cite{peng2018sfv} extracts human poses from monocular video, reconstructs reference motion trajectories, and then trains deep RL policies to follow these trajectories in simulation, demonstrating how perception-driven imitation can be combined with RL to achieve robust skill transfer across embodiments and environments.

\subsubsection{Domain Adaptation Hybrids}
\label{sec:domain_adaptation_hybrids}
Hybridization has also proven critical for addressing domain adaptation. Semantic Transfer Accelerated RL (STAR) \cite{pertsch2022cross} leverages demonstrations from different domains by encoding action sequences with a conditional VAE, pretraining low-level policies, and exploiting semantic priors to align states across tasks. By minimizing the KL divergence between semantically similar states and exploiting temporal context, STAR achieves efficient skill transfer across domains.

Another influential line of work uses unaligned video demonstrations. \cite{aytar2018playing}, for example, trained agents to play Atari by leveraging YouTube videos without direct domain alignment. Through self-supervised temporal distance classification and representation learning, their method enabled human-level performance in challenging exploration tasks, demonstrating that hybrid IL-RL models can generalize even from noisy, cross-domain data.

Residual learning further illustrates the utility of hybrid models, and works such as \cite{garcia2020physics} leveraged it to train a residual RL agent to refine noisy human hand poses for dexterous manipulation, with adversarial imitation ensuring corrections remain physically plausible. Similarly, \cite{stadie2017third} introduced third-person imitation learning, where RL combined with a GAN-based cost function recovery enabled imitation from videos with differing viewpoints and embodiments. More recent work \cite{huo2023efficient} extends this paradigm by leveraging task family priors and temporal abstractions extracted from demonstrations, alongside sampling-based Model Predictive Control (MPC) for safe trajectory generation.

Overall, hybrid approaches advance robot learning by combining IL’s efficiency, RL’s adaptability, and modern perception-control flexibility. This synergy enables robots to learn robust skills from a few demonstrations, transfer across domains and embodiments, and adapt policies online—bringing real-world deployment closer.

\begin{table*}[h!]
\centering
\footnotesize
\begin{tabular}{p{0.8cm} p{2.5cm} p{2.5cm} p{2.4cm} p{2.4cm} p{2.5cm}}
\toprule
\textbf{Method} & \textbf{Task Performance} & \textbf{Sample Efficiency} & \textbf{Compute Cost} & \textbf{Embodi. Handling} & \textbf{Pros (+) / Cons (-)} \\
\midrule

\cite{stadie2017third} & Succeeds on pointmass, reacher, and inverted pendulum (via 3rd-person demos) tasks & Efficient: no action/state alignment needed & Moderate: adversarial training + domain confusion & Explicit; learns domain-agnostic features & (+) Unsupervised, no action labels, handles domain gap; (-) Simple tasks only \\
\midrule

\cite{aytar2018playing} & Achieves and surpasses human-level in difficult Atari games (e.g., Montezuma) from video & Very efficient: single video demo is sufficient & High: deep self-supervised embeddings + RL training & Strong; domain-invariant video embeddings & (+) Solves sparse-reward/complex tasks, robust to visual gap; (-) Heavy compute, complex train \\
\midrule

\cite{peng2018sfv} & Learns high-fidelity dynamic skills (locomotion, acrobatics) & Efficient: uses public video data, minimal motion capture & High: deep pose estimation + RL with adaptive curriculum & Robust; physics-based policy handles noise & (+) Learns from unstructured video, retargets skills; (-) Needs good pose est., sim only \\
\midrule

\cite{garcia2020physics} & Improves success in dexterous VR manipulation and in-the-wild hand tracking & Needs mocap data for initial training, but less than full demo collection & Moderate (model-free hybrid RL + IL, hand pose estimation) & Residual policy corrects pose errors & (+) Physics-based, robust to estimation errors; (-) Needs mocap dataset for train \\
\midrule

\cite{pertsch2022cross} & Matches demo-accelerated RL for long-horizon kitchen tasks & Very efficient: $<$3 minutes human video enables long-horizon skill transfer & Moderate (semantic skill extraction, RL) & Robust; semantic imitation, cross-domain & (+) No in-domain demos, scalable, generalizes; (-) Needs offline skill extraction \\
\midrule

\cite{huo2023efficient} & High; enables one-shot fabric manipulation from video & Highly efficient: single demo + sim prior needed & Moderate: sim pretraining + MPC for policy optimization & No strict sim-to-real; scene-level alignment & (+) No risky real-world explore, efficient; (-) Focused on fabric, needs scene prior \\
\midrule

\cite{ramachandruni2020attentive} & Outperforms SOTA in pouring task imitation (lower error, fewer iterations) & Sample-efficient due to self-supervised and attention-guided feature learning & Moderate (CNN encoder + attention module + RL) & Attention-guided: view/background invariant & (+) Robust to clutter, learns focused features; (-) Needs multi-view, task-tuning \\
\bottomrule
\end{tabular}
\caption{Comparison of hybrid approaches for learning manipulation skills from human video}
\label{tab:hybrid_video_methods}
\end{table*}

Table \ref{tab:hybrid_video_methods} shows that hybrid models, integrating RL and IL, handle noisy demonstrations, sparse rewards, and domain adaptation well, excelling in cross-domain and viewpoint-variant tasks. However, they introduce design complexity and lack broad real-world validation. Overall, hybrid methods strike a balance between RL and IL strengths but remain best suited for simulation or moderately complex tasks.

\subsection{Multi-Modal Approaches}
\label{sec:multi_modal_methods}
\begin{figure*}[htbp] % htbp stand for "here", "top", "bottom", "page"
\includegraphics[scale=0.42]{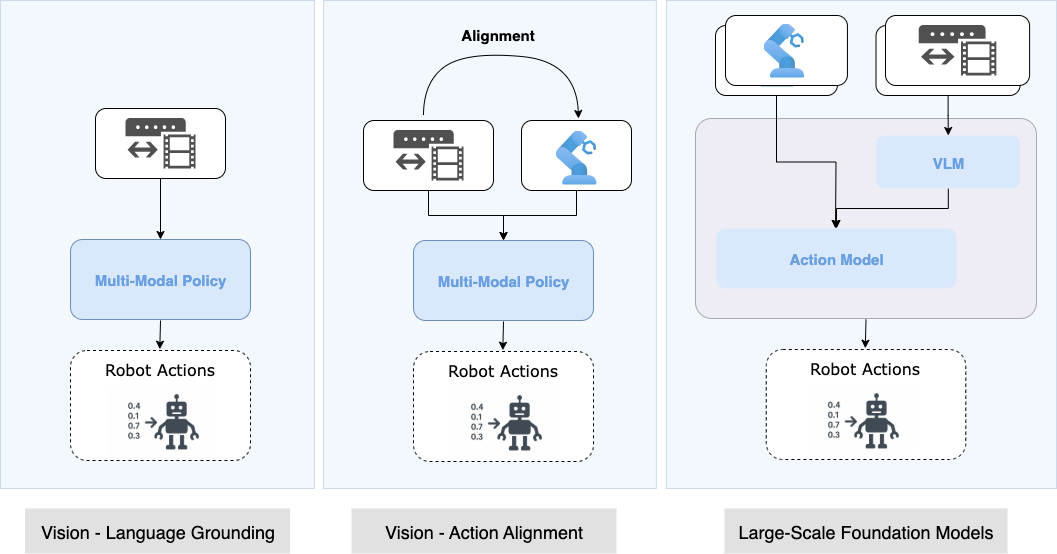} 
\caption{Evolution of multi-modal architectures in robot learning - Left: vision-language grounding, (Middle): vision-action alignment with shared embeddings, and Right: large-scale transformer-based foundation models (VLAs).}
\label{fig:multi_modal}
\end{figure*}

Robotic manipulation in unstructured environments requires the ability to interpret complex sensory signals and ground them in actionable policies. Relying on vision alone is often insufficient, as tasks typically involve abstract goals, contextual reasoning, or subtle cues that exceed purely visual perception. To address this, researchers have increasingly turned to multi-modal approaches (as shown in Figure \ref{fig:multi_modal}, where vision is combined with other input streams such as touch, proprioception, and, notably, natural language. By leveraging these complementary modalities, robots gain a richer and more holistic understanding of manipulation tasks, enabling them to follow nuanced instructions, reason about interactions, and generalize to unseen scenarios.

\subsubsection{Vision-Language Grounding}
\label{sec:vision_lang}

The earliest wave of vision-language methods focused on establishing a direct connection between natural language instructions and robotic action. For instance, \cite{shao2021concept2robot} fused natural language instructions with scene images to create task-specific embeddings that guided a policy network in generating robot trajectories. To enrich semantic understanding, the model leveraged a video-based action classifier trained on the Something-Something dataset \cite{goyal2017something}, aligning robot behavior with human demonstrations.

Before that, some researchers approached the topic from the standpoint of commonsense reasoning in robotics. \cite{jiang2020understanding} combined attention-based VLMs with ontology systems to represent manipulation concepts in time-independent semantic structures. Their introduction of the Robot Semantics Dataset and spatial attention mechanisms for action captioning laid the foundation for knowledge-graph reasoning in manipulation.

There has also been progress with methods that linked low-level perception to higher-level linguistic descriptions. For example, \cite{yang2019learning} proposed a dual-model architecture: a grasp detection network (GNet) that computed object grasps and a captioning network (CNet) that translated demonstration videos into commands. Here, language was not only a means of communication but also a scaffold for structuring perception and action.  

Building on these ideas, works such as Watch and Act \cite{10064325} introduced pipelines where video captioning and robot planning were tightly coupled. Demonstration videos were first converted into textual instructions, which were then grounded in visual perception modules (e.g., segmentation) and executed through RL-based controllers. This marked a shift toward systems capable of seamless transitions from naturalistic demonstrations to executable actions.  

Generalization has also been an active focus in vision-language learning. \cite{zhang2019zero}, for instance, enabled robots to perform zero-shot imitation of both single-agent and collaborative tasks from YouTube videos. Instead of generating commands, their framework constructed action grammars and action trees, providing a structured yet flexible approach to representing novel behaviors. These advances highlight the progression from basic grounding of instructions to more scalable frameworks that support open-ended learning.

\subsubsection{Vision-Action Alignment via Multi-Modal Representations}
\label{sec:large_foundation_models}

While grounding language in perception was a critical first step, recent advances extend beyond mapping words to actions toward building unified multimodal representations. These approaches jointly embed videos, states, and textual instructions into shared spaces for reasoning and control \cite{zitkovich2023rt,padalkar2023open}. This integration moves the field from task-specific pipelines to generalizable frameworks that bridge perception and action more seamlessly.

Early attempts at such alignment explored direct translation of demonstrations into robot instructions. \cite{nguyen2018translating} employed CNNs and RNNs to convert visual features into grammar-free command descriptions, demonstrating how semantic comprehension could augment traditional imitation learning. Subsequent approaches like Vid2Robot \cite{jain2024vid2robot} refined this idea by using cross-attention to align video prompts with a robot’s state, with contrastive losses ensuring robust representation learning. These works underscored the value of multi-modal alignment in supporting long-horizon reasoning and motion transfer.

The ability to predict future dynamics further broadened the scope of these models. \cite{yuan2024general} proposed predicting “general flow” 3D trajectories of object points from RGB-D video and language input, enabling skill transfer across embodiments and morphologies. Similarly, \cite{dasari2021transformers} leveraged transformers for one-shot imitation from videos, introducing an inverse dynamics loss to stabilize self-attention and improve policy adaptation. Together, these models illustrate a progression from mapping demonstrations to generalizable trajectory prediction.

With internet-scale video data becoming available, generative architectures have pushed multi-modal learning into broader domains. \cite{bharadhwaj2023zero} trained an image transformer with a conditional variational autoencoder (C-VAE) to anticipate human actions and object interactions from diverse online videos. The resulting model achieved zero-shot transfer to novel lab settings, demonstrating the potential of pretraining on open-world data. Building on this principle, PLanning-EXecution (PLEX) \cite{thomas2023plex} introduced a planner-executor transformer framework that separates high-level activity sequencing from low-level action execution, supporting multi-task generalization even in low-data regimes.  

Taken together, these advances chart a clear trajectory: from early instruction-to-action systems to transformer-based architectures unifying vision, language, and action, and finally to large-scale pretraining for robust generalization. Multi-modal representations have become a cornerstone of next-generation robotic intelligence, enabling grounding, abstraction, and synthesis for versatile real-world manipulation.

\begin{table*}[h!]
\centering
\footnotesize
\begin{tabular}{p{0.8cm} p{2.5cm} p{2.5cm} p{2.4cm} p{2.4cm} p{2.5cm}}
\toprule
\textbf{Method} & \textbf{Task Performance} & \textbf{Sample Efficiency} & \textbf{Compute Cost} & \textbf{Embodi. Handling} & \textbf{Pros (+) / Cons (-)} \\
\midrule

\cite{nguyen2018translating} & Strong hierarchical task generalization and adaptation & Efficient due to hierarchical reuse & Moderate (modular network execution) & Partial handling via sub-program abstraction & (+) Good  Compositionality and generality; (-) Needs task sketch annotations \\
\midrule

\cite{yang2019learning} & Good for long-horizon tasks & Needs many demos & High due to GCNs and temporal reasoning & Not directly addressed & (+) Task-structure aware; (-) High memory/sample cost \\
\midrule

\cite{zhang2019zero} & Good zero-shot on unseen tasks due to pretraining with human video) & Very efficient; human video only for most of training & Moderate-high & Object-centric/temporal cues bridge gap & (+) Pure visual imitation; (-) Needs temporal coherence \\
\midrule

\cite{jiang2020understanding} & Good for vision-language manipulation & Relatively data-efficient due to CLIP feature reuse & Moderate-high & Indirect via pretrained Vision-Language features & (+) Language generalizes; (-) Fails on vague instructions \\
\midrule

\cite{jain2024vid2robot} & High; 20\% improvement over BC-Z on real robots tasks & Data-efficient; due to paired video/trajectory data & High (large transformer with cross-attention over videos) & Contrastive loss/data pairing bridge gap & (+) Imitate from human demos; (-) Data collection bottleneck \\
\midrule

\cite{shao2021concept2robot} & High; learns 78 tasks from human videos; strong simulated generalization across instructions/scenes & Efficient: self-supervised RL, no teleop demos & Moderate (RL + supervised imitation; video classifier training) & Video reward model bridges sim2real & (+) Diverse skills from unstructured video; (-) No real-robot validation, open-loop \\
\midrule

\cite{10064325} & High; validated on 24 objects, 8 actions, and real robot tasks & Moderate: relies on synthetic data for segmentation; video demos for policy learning & Moderate (captioning, segmentation, RL-based affordance) & Captioning/modular policy reduce gap & (+) No expert action data, real-world diversity; (-) Pipeline complexity, caption accuracy dep. \\
\midrule

\cite{yuan2024general} & 81\% success across 18 real-world tasks & No robot data; trained on human videos only & Low (simple policy; training cost in flow model) & Very low gap via flow affordance & (+) Embodiment-agnostic; (-) Less expressive than trajectories \\
\midrule

\cite{dasari2021transformers} & High; 2x better than prior pick-place baselines in sim (16 tasks) & Moderate; needs 1600 demo-context pairs across tasks; one-shot at test time & Moderate; Transformer-based encoder-decoder with ResNet and multi-loss training (BC + inverse + keypoint loss) & Partial; visual transfer across Sawyer/Panda in sim & (+) Strong generalization via attention; (-) Not real-world tested, needs good alignment \\
\midrule

\cite{bharadhwaj2023zero} & Good; 50\% (unconditioned) and 37\% (goal-conditioned) real robot success & High efficiency; uses internet videos only, no robot data; needs hand pose estimates & Moderate; Transformer-CVAE architecture with inverse kinematics controller. Requires camera calibration & Strong; hand trajectory mapping with camera transforms, IK & (+) No lab data, real-world tested; (-) Pose estimate noise, limited precision \\
\midrule

\cite{thomas2023plex} & High; SOTA on Meta-World/Robosuite (100\% Lift, \textasciitilde{88}\% insert) & Very high; pretrains on 4500 video demos + 375 visuomotor, fine-tune with as few as 10 demos & High; Ddual Transformer (planner + executor) architecture with positional encoding and inverse dynamics training & Moderate; generalizes tasks, but assumes fixed robot morphology & (+) Multi-modal, few-shot; (-) High compute, needs curated data splits \\
\bottomrule
\end{tabular}
\caption{Comparative analysis of vision-language grounding and vision-action alignment approaches for learning manipulation skills from human video}
\label{tab:multi_modal_video_methods}
\end{table*}

Table \ref{tab:multi_modal_video_methods} highlights the strong ability of these methods to integrate vision, language, and sometimes other modalities, supporting flexible, language-driven, and zero-shot manipulation. They excel at generalization, especially for long-horizon and open-vocabulary tasks, but face challenges in model complexity, data demands, and training cost. Thus, while powerful and general, multi-modal approaches remain constrained by the need for rich datasets and significant computational resources.

\subsubsection{Multi-Modal Transformers and Large-Scale Foundation Models}
\label{sec:foundation_model}

While early multi-modal approaches established the feasibility of combining perception, action, and language, their scalability remained limited. The arrival of large-scale transformer architectures and language-conditioned policy models has since transformed the landscape, providing the representational capacity and flexibility needed to tackle long-horizon, multi-task robot learning. These advances are driven by a common intuition: that transformer-based models, pretrained on diverse multimodal data, can capture the temporal, semantic, and structural regularities necessary for robust manipulation policies.

Initial breakthroughs demonstrated the power of hierarchical attention mechanisms for grounding multimodal signals. Hierarchical Universal Language Conditioned Policies (HULC) \cite{mees2022matters}, for example, employed a hierarchical transformer encoder with contrastive alignment of video and language embeddings to support long-horizon manipulation. Evaluated on Composing Actions from Language and Vision (CALVIN) \cite{mees2022calvin}, HULC achieved strong generalization across tasks, marking one of the first demonstrations that multimodal transformers could scale beyond narrow, single-task pipelines. Building on this foundation, models such as VisuoMotor Attention (VIMA) \cite{jiang2022vima} introduced transformer-based architectures that process multimodal prompts composed of visual and textual tokens. By incorporating pretrained language models into an encoder-decoder system, VIMA was able to perform data-efficient policy learning across a wide range of manipulation tasks, showing how prompting can unify task specification and execution. Similarly, DigKnow \cite{chen2023human} leveraged LLMs to extract layered knowledge from scene graphs in human demonstration videos, enabling retrieval and correction mechanisms that improve generalization to novel task instances.

A major shift occurred with the adoption of generative pretraining paradigms. GR-1 \cite{wu2023unleashing}, for instance, unified instructions, video observations, and robot states into a single predictive architecture. By pretraining on massive video datasets and fine-tuning on robot data, GR-1 achieved state-of-the-art results on challenging manipulation benchmarks. Similarly, Video-based Policy learning framework via Discrete Diffusion (VPDD) \cite{he2024large} demonstrated how discrete diffusion models could compress video data into latent tokens, predict future video dynamics, and then fine-tune with limited robot-labeled data. These approaches established video pretraining as a key enabler of scalable robot learning.

Recent VLA models have pushed these ideas further by leveraging hundreds of millions of internet videos to endow robots with broad physical priors and temporal reasoning. For example, GR-2 \cite{cheang2024gr} was pretrained on 38 million videos to learn conditional temporal prediction, then fine-tuned on robot demonstrations to achieve state-of-the-art multi-task performance (as shown in Figure \ref{fig:vla_arch}). GROOT N1 \cite{bjorck2025gr00t} extended this idea with a latent action codebook, enabling even action-less human videos to be repurposed as robot data. By generating synthetic “neural trajectories” through video generation models, it amplified its training corpus dramatically and, with a diffusion transformer, achieved robust multimodal control across diverse robots. Other models have explored new mechanisms for integrating human video data into robot learning. CoT-VLA \cite{zhao2025cot} introduced a visual chain-of-thought framework, predicting subgoal images before producing corresponding actions. This allowed training on large-scale human activity datasets such as Epic-Kitchens, despite the lack of action annotations, improving long-horizon reasoning. Gemini Robotics \cite{team2025gemini}, meanwhile, extended Gemini 2.0 with an Embodied Reasoning (ER) layer. By training on massive web video corpora, it acquired structured perception primitives, such as 3D object detection, grasp prediction, and trajectory reasoning, that could be adapted to novel tasks with only in-context prompting or minimal fine-tuning.

\begin{table*}[h!]
\centering
\footnotesize
\begin{tabular}{p{0.8cm} p{2.5cm} p{2.5cm} p{2.4cm} p{2.4cm} p{2.5cm}}
\toprule
\textbf{Method} & \textbf{Task Performance} & \textbf{Sample Efficiency} & \textbf{Compute Cost} & \textbf{Embodi. Handling} & \textbf{Pros (+) / Cons (-)} \\
\midrule

\cite{mees2022matters} & SoTA on CALVIN: strong multi-stage, long-horizon manipulation & Moderate; leverages relabeled play/language data & Moderate-high (multimodal transformer + hierarchical structure) & Grounded language/vision & (+) Many tasks/flexible; (-) Model/training complexity \\
\midrule

\cite{jiang2022vima} & SoTA in visuo-motor multimodal tasks & Highly data-efficient via large-scale pretraining & High (transformer + large multimodal model) & Good; 3D scene reduces sim2real gap & (+) Strong zero-shot; (-) Expensive train/deploy \\
\midrule

\cite{chen2023human} & High; improves planning \& execution & Efficient (retrieves knowledge instead of retraining) & Moderate (scene graph, LLM, simple policy) & Hierarchical narrows gap & (+) No retrain; (-) Needs scene graph/LLM reliability \\
\midrule

\cite{wu2023unleashing} & SoTA on CALVIN and real robot tasks (94.9\% success) & Very efficient (finetunes on 10\% robot data) & High (GPT-style video transformer) & Good; finetunes after pretraining & (+) Best generalization/few-shot; (-) High model/training cost \\
\midrule

\cite{he2024large} & SoTA on Meta-World, RLBench (both seen/unseen tasks) & High; learns from large unlabeled video, and requires few labeled demos & High (VQ-VAE, discrete diffusion, multi-stage training) & Visual token space bridges videos & (+) Leverages internet-scale data; (-) Compute heavy; limited real-robot \\
\midrule

\cite{bjorck2025gr00t} & High; strong performance across simulation benchmarks & High; trained on diverse real-robot trajectories, human videos, and synthetic data & High; dual-system architecture with diffusion transformer & Excellent; tightly coupled VLA modules & (+) Open-source; (+) High data efficiency; (-) Complex model architecture \\
\midrule

\cite{cheang2024gr} & High; capable of completing 105 manipulation tasks with high success rate & High; pre-trained on 38 million text-video data & High; utilizes large multimodal model architecture & Good; supports various robot embodiments & (+) Strong generalization; (-) High compute requirements \\
\midrule

\cite{zhao2025cot} & High; outperforms state-of-the-art VLA models in real-world manipulation tasks & High; incorporates visual chain-of-thought reasoning for efficient task execution & Moderate; utilizes 7B parameter model & Good; trained on robot demonstrations and action-less videos & (+) Enhanced reasoning capabilities; (-) Requires fine-tuning for new tasks \\
\midrule

\cite{team2025gemini} & High; excels in dexterous manipulation tasks & High; optimized for on-device processing with minimal computational resources & Moderate; designed for local inference on robots & Excellent; supports various robot embodiments & (+) On-device operation; (+) Adaptable to new tasks; (-) Limited to bi-arm robots \\
\bottomrule

\end{tabular}
\caption{Comparison of multi-modal transformers and large-scale foundation models for learning manipulation skills from human video}
\label{tab:large_foundation_models}
\end{table*}

Collectively, these works trace a clear trajectory: from hierarchical multimodal transformers, to prompt-based frameworks, to generative pretraining, and finally to large-scale VLA models that leverage internet video at unprecedented scales. The unifying theme is the shift from narrow, demonstration-driven learning toward architectures that internalize broad physical common sense, temporal dynamics, and general reasoning. These advances position multimodal foundation models not just as tools for policy learning, but as platforms for embodied intelligence with the capacity to adapt fluidly to new environments and tasks. Despite the many gains won by these models, Table \ref{tab:large_foundation_models} shows their comparative extreme compute and data requirements make them unsuitable for low-resource settings.

\section{Comparative Analysis Between the Approaches}
\label{sec:comparative_analysis}

To provide a high-level synthesis of the literature, we present in Table \ref{tab:macro_comparison} a macro-level comparison of the primary methodological subgroups discussed in this survey. While previous sections have detailed the specific technical approaches within each subgroup, this section aims to capture their overarching strengths, common pitfalls, bottlenecks, and representative applications.  

The chosen metrics' main advantages, disadvantages, key bottlenecks, and reported applications reflect the most salient aspects influencing the applicability and impact of each subgroup in practical robot learning settings.

\begin{table*}[h!]
\centering
\footnotesize
\begin{tabular}{p{1.8cm} p{3.0cm} p{3.0cm} p{2.4cm} p{3.0cm}}
\toprule
\textbf{Category} & \textbf{Advantages} & \textbf{Disadvantages} & \textbf{Key Bottlenecks} & \textbf{Representative Applications} \\
\midrule
\textbf{Foundational Perception \& Representation Methods} & Interpretable representations; leverages large unlabeled datasets; real-time capable; enables structured motion capture; bridges visual domain gaps; adapts across environments & Limited to holistic features; dependent on estimation accuracy; translation artifacts; requires diverse contextual data; vulnerable to domain shift & Embodiment transfer, pose estimation robustness, visual realism vs. task fidelity, context generalization & Hand-object interaction, dexterous grasping, human-to-robot transfer, cross-domain manipulation, assembly tasks \\
\midrule
\textbf{Reinforcement Learning Approaches} & Powerful for long-horizon tasks; handles complex manipulation; enables compositional learning; grounds policies in structured representations; supports multi-task generalization & High computational cost; complex architecture design; sensitive to reward engineering; challenging hierarchical decomposition; requires extensive interaction data & Feature representation quality, reward signal design, task decomposition, hierarchical learning stability & Object manipulation with dense rewards, multi-step assembly, hierarchical manipulation, tool use learning, compositional skill learning \\
\midrule
\textbf{Imitation Learning Approaches} & Direct demonstration use; sample efficient; rapid skill acquisition; enables few-shot generalization; works with raw video data; leverages web-scale demonstrations & Distribution mismatch issues; requires extensive meta-training; embodiment mismatch challenges; sensitive to demonstration quality; limited generalization beyond training & Demonstration coverage, meta-learning stability, domain gap bridging, embodiment alignment, visual domain shift & One-shot imitation, few-shot manipulation, rapid task adaptation, human video imitation, cross-embodiment transfer \\
\midrule
\textbf{Hybrid Approaches} & Combines RL/IL strengths; robust to demonstration noise; addresses cross-domain challenges; enables online refinement; flexible adaptation mechanisms & Increased design complexity; challenging balance optimization; complex pipeline design; limited real-world validation; task-specific solutions & Integration complexity, multi-objective learning, cross-domain alignment, multi-component integration, real-world deployment & Noisy demonstration learning, online policy refinement, cross-domain skill transfer, adaptive imitation, robust policy learning \\
\midrule
\textbf{Multi-Modal Approaches} & Natural language task specification; unified representations; internet-scale pretraining; broad physical common sense; supports in-context learning; enables zero-shot generalization & Requires paired multimodal data; high model complexity; extremely high compute requirements; sensitive to language ambiguity; deployment challenges & Language-vision alignment, multimodal representation learning, computational scalability, data curation, model interpretability & Language-guided manipulation, trajectory prediction, general manipulation policies, zero-shot task execution, embodied reasoning \\
\bottomrule
\end{tabular}
\caption{Macro-level comparative analysis of video-based robot learning approaches.}
\label{tab:macro_comparison}
\end{table*}

The comparative analysis reveals a field grappling with fundamental tensions between capability and practicality. At the foundational level, perception approaches illustrate this tension clearly: CNN-based methods offer computational efficiency but sacrifice fine-grained understanding, while pose/keypoint detection provides structured representations at the cost of robustness to estimation failures. This same pattern extends to domain bridging, where image translation prioritizes visual alignment over action precision, and context translation emphasizes environmental adaptability over training efficiency.

These trade-offs become more pronounced as approaches increase in sophistication. Reinforcement learning methods exemplify this progression, from computationally expensive visual RL to hierarchical frameworks that promise compositional learning but introduce complex decomposition challenges. Similarly, imitation learning has evolved from direct behavioral cloning toward meta-learning approaches that achieve impressive few-shot capabilities but require extensive pretraining regimes that paradoxically reduce their practical accessibility.

The field's response to these limitations through hybrid and multi-modal approaches reveals both promise and deeper challenges. Hybrid methods acknowledge that single paradigms are insufficient, yet their attempt to combine multiple learning objectives often creates complex, task-specific solutions that resist broader generalization. Multi-modal approaches, particularly large-scale foundation models, represent the current apex of capability but exacerbate the accessibility problem by requiring computational resources beyond most practitioners' reach.

This evolution exposes a central paradox: the most capable methods are becoming increasingly inaccessible, while practical approaches face fundamental limitations in handling the embodiment transfer problem that persists across all categories. The field has not solved this core challenge but rather developed increasingly sophisticated ways to work around it, often at the cost of practical deployability.

The implications are clear: method selection cannot be driven solely by theoretical performance but must account for the specific constraints of computational resources, data availability, and deployment requirements. Rather than seeking universal solutions, the field is converging on a recognition that different approaches occupy distinct positions in a multi-dimensional trade-off space, making the matching of methods to specific scenarios the critical skill for practical robot learning applications.

\section{Open-Source Tools for Video-Based Robot Manipulation Learning}
\label{sec:open_source_tools}
In this section, we discuss and provide an overview of open-source implementations, frameworks, tools, and datasets that constitute the foundation of modern video-based robot manipulation learning. The resources cataloged herein in \cref{tab:self_supervised,tab:3d_modeling,tab:affordance,tab:vla,tab:datasets,tab:libraries}, span the entire learning pipeline, from foundational visual representation models that serve as the perceptual backbone of a system, to sophisticated end-to-end VLA policies that map raw pixels and natural language commands directly to robot motor commands.

%================================================================%
% Table 1: Foundational Visual Representation Learning
%================================================================%
\begin{table*}[h!]
\centering

\begin{tabularx}{\textwidth}{@{} p{2.8cm} p{2.8cm} p{5.5cm} p{2.5cm} @{}}
\toprule
\textbf{Resource Name} & \textbf{Primary Function} & \textbf{Key Features} & \textbf{Open-Source Link} \\
\midrule
Time-Contrastive Networks (TCN) \cite{sermanet2018time} & Self-Supervised Representation Learning & 
    \begin{itemize}[nosep, leftmargin=*]
        \item Learns viewpoint-invariant features from multi-view video
        \item Uses a time-contrastive triplet loss
        \item Focuses on temporal dynamics over static appearance
    \end{itemize} & \url{https://github.com/kekeblom/tcn} \\
\addlinespace
Spatiotemporal Contrastive Video Representation Learning (CVRL) \cite{qian2021spatiotemporal} & Self-Supervised Representation Learning & 
    \begin{itemize}[nosep, leftmargin=*]
        \item Learns spatiotemporal features via contrastive loss
        \item Employs temporally consistent spatial augmentations
        \item Outperforms ImageNet pre-training on video tasks
    \end{itemize} & \url{https://github.com/tensorflow/models/tree/master/official/} \\
\addlinespace
Dense Predictive Coding (DPC) \cite{han2019video} & Self-Supervised Representation Learning & 
    \begin{itemize}[nosep, leftmargin=*]
        \item Predicts future latent representations, not pixels
        \item Learns dense spatio-temporal block embeddings
        \item Uses curriculum learning to predict further in time
    \end{itemize} & \url{https://github.com/TengdaHan/DPC} \\
\addlinespace
Masked Visual Pre-training (MVP) \cite{xiao2022masked} & Self-Supervised Representation Learning & 
    \begin{itemize}[nosep, leftmargin=*]
        \item Extends Masked Autoencoders (MAE) to robotics
        \item Pre-trains on large image/video datasets
        \item Frozen encoder serves as perception module for control
    \end{itemize} & \url{https://github.com/ir413/mvp} \\
\addlinespace
R3M \cite{nair2022r3m} & Universal Visual Representation & 
    \begin{itemize}[nosep, leftmargin=*]
        \item Pre-trained on Ego4D human video dataset
        \item Combines time-contrastive and video-language learning
        \item Serves as a frozen perception module for manipulation
    \end{itemize} & \url{https://sites.google.com/view/robot-r3m} \\
\bottomrule
\end{tabularx}
\caption{Foundational Visual Representation Learning Resources}
\label{tab:self_supervised}
\end{table*}

%================================================================%
% Table 2: 3D Hand & Body Modeling
%================================================================%
\begin{table*}[h!]
\centering
\begin{tabularx}{\textwidth}{@{} p{2.8cm} p{2.8cm} p{5.5cm} p{2.5cm} @{}}
\toprule
\textbf{Resource Name} & \textbf{Primary Function} & \textbf{Key Features} & \textbf{Open-Source Link} \\
\midrule
SMPL \cite{SMPL:2015} & Parametric 3D Body Model & 
    \begin{itemize}[nosep, leftmargin=*]
        \item Learned from thousands of 3D body scans
        \item Low-dimensional shape and pose parameterization
        \item Compatible with standard graphics engines
    \end{itemize} & \url{https://smpl.is.tue.mpg.de/} \\
\addlinespace
MANO \cite{romero2022embodied} & Parametric 3D Hand Model & 
    \begin{itemize}[nosep, leftmargin=*]
        \item Specialized parametric model for the human hand
        \item Integrates with SMPL to form SMPL+H
        \item Enables detailed, articulated hand modeling
    \end{itemize} & \url{https://mano.is.tue.mpg.de/} \\
\addlinespace
FrankMocap \cite{rong2020frankmocap} & Monocular Motion Capture System & 
    \begin{itemize}[nosep, leftmargin=*]
        \item Real-time 3D hand and body motion capture from single RGB video
        \item Leverages SMPL-X for unified parametric output
        \item Enables data extraction from in-the-wild videos
    \end{itemize} & \url{https://github.com/facebookresearch/frankmocap} \\
\bottomrule
\end{tabularx}
\caption{3D Hand \& Body Modeling Resources}
\label{tab:3d_modeling}
\end{table*}

%================================================================%
% Table 3: Affordance & Interaction
%================================================================%
\begin{table*}[h!]
\centering
\begin{tabularx}{\textwidth}{@{} p{2.8cm} p{2.8cm} p{5.5cm} p{2.5cm} @{}}
\toprule
\textbf{Resource Name} & \textbf{Primary Function} & \textbf{Key Features} & \textbf{Open-Source Link} \\
\midrule
AffordanceNet \cite{do2018affordancenet} & Affordance Detection & 
    \begin{itemize}[nosep, leftmargin=*]
        \item End-to-end object and affordance detection from RGB-D
        \item Uses a multi-task, two-branch architecture
        \item Segments pixels into functional categories (e.g., "grasp")
    \end{itemize} & \url{https://github.com/wliu88/affordance_net} \\
\addlinespace
Demo2Vec \cite{fang2018demo2vec} & Affordance Reasoning & 
    \begin{itemize}[nosep, leftmargin=*]
        \item Learns video embeddings to reason about affordances
        \item Predicts interaction heatmaps and action labels on a target image
        \item Trained on YouTube product review videos
    \end{itemize} & \url{https://sites.google.com/view/demo2vec/} \\
\addlinespace
Vision-Robotics Bridge (VRB) \cite{bahl2023affordances} & Affordance Grounding & 
    \begin{itemize}[nosep, leftmargin=*]
        \item Learns agent-agnostic affordances from human videos
        \item Predicts contact heatmaps and post-contact trajectories
        \item Integrates with multiple robot learning paradigms
    \end{itemize} & \url{https://robo-affordances.github.io/} \\
\bottomrule
\end{tabularx}
\caption{Affordance \& Interaction Resources}
\label{tab:affordance}
\end{table*}

%================================================================%
% Table 4: VLA Policies
%================================================================%
\begin{table*}[h!]
\centering
\begin{tabularx}{\textwidth}{@{} p{2.8cm} p{2.8cm} p{5.5cm} p{2.5cm} @{}}
\toprule
\textbf{Resource Name} & \textbf{Primary Function} & \textbf{Key Features} & \textbf{Open-Source Link} \\
\midrule
RT-1 \cite{brohan2022rt} & Transformer-Based Policy & 
    \begin{itemize}[nosep, leftmargin=*]
        \item End-to-end Transformer for multi-task control
        \item Tokenizes images, language, and actions
        \item Trained on 130k+ real-world robot trajectories
    \end{itemize} & \url{https://github.com/google-research/robotics_transformer} \\
\addlinespace
RT-2 \cite{zitkovich2023rt} & Vision-Language-Action Model & 
    \begin{itemize}[nosep, leftmargin=*]
        \item Fine-tunes web-scale VLMs for robotic control
        \item Represents robot actions as text tokens
        \item Transfers semantic knowledge from web data to robotics
    \end{itemize} & \url{https://github.com/kyegomez/RT-2} \\
\addlinespace
CLIPort \cite{shridhar2022cliport} & Language-Conditioned Imitation & 
    \begin{itemize}[nosep, leftmargin=*]
        \item Two-stream architecture: semantic (CLIP) and spatial (Transporter)
        \item Combines "what" (language) and "where" (affordance)
        \item Generalizes to unseen objects without explicit detectors
    \end{itemize} & \url{https://cliport.github.io/} \\
\addlinespace
VIMA \cite{jiang2022vima} & Multimodal Prompting Agent & 
    \begin{itemize}[nosep, leftmargin=*]
        \item Generalist Transformer agent for diverse tasks
        \item Unifies task specification via multimodal prompts (text + vision)
        \item Trained on 600k+ expert trajectories in VIMA-Bench
    \end{itemize} & \url{https://vimalabs.github.io/} \\
\bottomrule
\end{tabularx}
\caption{Vision-Language-Action (VLA) Policies}
\label{tab:vla}
\end{table*}

%================================================================%
% Table 5: Datasets & Simulators
%================================================================%
\begin{table*}[h!]
\centering
\begin{tabularx}{\textwidth}{@{} p{2.8cm} p{2.8cm} p{5.5cm} p{2.5cm} @{}}
\toprule
\textbf{Resource Name} & \textbf{Primary Function} & \textbf{Key Features} & \textbf{Open-Source Link} \\
\midrule
Open X-Embodiment (OXE) \cite{padalkar2023open} & Cross-Embodiment Dataset & 
    \begin{itemize}[nosep, leftmargin=*]
        \item Unifies 60+ datasets from 22 robot embodiments
        \item 1M+ real robot trajectories in standardized RLDS format
        \item Enables training of generalist "X-robot" policies
    \end{itemize} & \url{https://github.com/google-deepmind/open_x_embodiment} \\
\addlinespace
DROID \cite{khazatsky2024droid} & In-the-Wild Manipulation Dataset & 
    \begin{itemize}[nosep, leftmargin=*]
        \item 76k+ trajectories from 50+ global users
        \item Collected in 564 diverse, unstructured scenes
        \item Designed for studying real-world generalization
    \end{itemize} & \url{https://droid-dataset.github.io/} \\
\addlinespace
BridgeData V2 \cite{walke2023bridgedata} & Multi-Task Manipulation Dataset & 
    \begin{itemize}[nosep, leftmargin=*]
        \item 60k+ trajectories on a low-cost WidowX robot
        \item Spans 24 environments and 13 skills
        \item Includes language and goal-image conditioning
    \end{itemize} & \url{https://rail-berkeley.github.io/bridgedata/} \\
\addlinespace
RoboNet \cite{dasari2019robonet} & Multi-Robot Video Dataset & 
    \begin{itemize}[nosep, leftmargin=*]
        \item 15M video frames from 7 different robot platforms
        \item Early large-scale effort to share robotic experience
        \item Designed for learning generalizable vision-based models
    \end{itemize} & \url{https://www.robonet.wiki/} \\
\addlinespace
RH20T \cite{fang2023rh20t} & Contact-Rich Manipulation Dataset & 
    \begin{itemize}[nosep, leftmargin=*]
        \item 110k+ sequences focusing on contact-rich skills
        \item Rich multi-modal data (vision, force, audio, action)
        \item Includes paired human demonstration for each robot sequence
    \end{itemize} & \url{https://rh20t.github.io/} \\
\addlinespace
Isaac Sim \cite{NVIDIA_Isaac_Sim} & GPU-Based Physics Application & 
    \begin{itemize}[nosep, leftmargin=*]
        \item High-performance simulator for massively parallel RL
        \item Enables training policies directly on GPU
        \item Developed by NVIDIA
    \end{itemize} & \url{https://developer.nvidia.com/isaac/sim} \\
\addlinespace
MuJoCo \cite{6386109} & Physics Engine & 
    \begin{itemize}[nosep, leftmargin=*]
        \item Fast and accurate physics simulation
        \item Widely used for robotics, biomechanics, and RL research
        \item Features optimization-based contact dynamics
    \end{itemize} & \url{https://mujoco.org/} \\
\addlinespace
PyBullet \cite{coumans2021} & Physics Simulator & 
    \begin{itemize}[nosep, leftmargin=*]
        \item Python module for the Bullet physics engine
        \item Provides robotics simulation and RL environments
        \item Open-source and widely accessible
    \end{itemize} & \url{https://pybullet.org/} \\
\bottomrule
\end{tabularx}
\caption{Datasets \& Simulators}
\label{tab:datasets}
\end{table*}

%================================================================%
% Table 6: Core Libraries
%================================================================%
\begin{table*}[h!]
\centering
\begin{tabularx}{\textwidth}{@{} p{2.8cm} p{2.8cm} p{5.5cm} p{2.5cm} @{}}
\toprule
\textbf{Resource Name} & \textbf{Primary Function} & \textbf{Key Features} & \textbf{Open-Source Link} \\
\midrule
OpenCV \cite{OpenCV2010} & Computer Vision Library & 
    \begin{itemize}[nosep, leftmargin=*]
        \item De facto standard library for real-time image/video processing
        \item Provides a vast suite of foundational CV algorithms
        \item Underpins most vision-based robotics research
    \end{itemize} & \url{https://opencv.org/} \\
\addlinespace
MediaPipe \cite{48292} & Perception Pipeline Framework & 
    \begin{itemize}[nosep, leftmargin=*]
        \item Cross-platform framework for applied ML pipelines
        \item Offers pre-trained models for pose, hand, and face tracking
        \item Used as off-the-shelf perception components
    \end{itemize} & \url{https://developers.google.com/mediapipe} \\
\addlinespace
OpenPose \cite{cao2019openpose} & 2D Pose Estimation & 
    \begin{itemize}[nosep, leftmargin=*]
        \item Real-time multi-person 2D keypoint detection
        \item Detects body, hand, foot, and facial keypoints
        \item Widely used for human activity analysis
    \end{itemize} & \url{https://github.com/CMU-Perceptual-Computing-Lab/openpose} \\
\bottomrule
\end{tabularx}
\caption{Core Libraries}
\label{tab:libraries}
\end{table*}

\section{Challenges}
\label{sec:challenges}

Learning from video demonstrations presents several persistent challenges that cut across all major methodological subgroups. While remarkable progress has been made, limitations remain at both the data and model levels, and many of these challenges directly constrain the practical impact of state-of-the-art methods. This section identifies and discusses six key challenges in learning from video demonstrations, expanding on: (1) data availability and annotation, (2) domain shift and embodiment gap, (3) computational cost and scalability of learning architectures and resources, (4) model sample efficiency, (5) evaluation and benchmarking, and (6) causal reasoning and policy abstraction.

\subsection{Data Availability and Annotation}
\label{sec:data_availability}

The performance and generalization of video-based robot learning models remain highly dependent on the availability, diversity, and quality of demonstration data. As highlighted in our analysis, many approaches, especially feature extraction and imitation learning struggle when exposed to unfamiliar states or objects not seen during training. Large, balanced datasets are rare, and most existing datasets (e.g., EPIC-Kitchens \cite{damen2018scaling}, Something-Something \cite{goyal2017something}) are often domain-specific, unbalanced, or require expensive annotation. Specialized datasets such as Penn Action \cite{6751390}, HMDB51 \cite{kuehne2011hmdb}, and MPII \cite{andriluka14cvpr} are often limited in their diversity or designed for narrow, non-robotic tasks. Furthermore, several methods (e.g., Demo2Vec \cite{fang2018demo2vec}, ViP \cite{chane2023learning}) depend on expert-labeled demonstrations, further limiting their scalability in real-world scenarios. The community increasingly turns to scalable alternatives, such as leveraging uncurated internet videos, weak supervision, or self-supervised objectives, to address these data bottlenecks, but progress is ongoing.

\subsection{Domain Shift and Embodiment Gap}
\label{sec:domain_shift}

Domain shift remains a fundamental challenge across all subgroups. The disparity between human and robot domains, the so-called embodiment gap, often hinders the direct transfer of skills, as the visual appearance, dynamics, and even the action spaces differ substantially. Although some methods leverage domain-invariant or domain-adaptive features, as seen in hybrid, multi-modal, and image/context translation approaches, the problem is far from solved. Translation artifacts, imperfect pose estimation, and misaligned action representations can severely limit sim-to-real or human-to-robot transfer. Approaches like adversarial learning, keypoint-based transfer, and CycleGANs offer partial solutions, but robust, generalizable transfer remains elusive.

\subsection{Computational Cost}
\label{sec:compute_cost}

Computational cost is an increasingly pressing concern, especially with the adoption of large-scale, multi-modal, and transformer-based architectures. While these models enable impressive performance, their high computational and memory requirements can limit real-world deployment, especially on resource-constrained robotic platforms. Many state-of-the-art methods (e.g., multi-modal and large-scale models \ref{sec:multi_modal_methods}) demand substantial GPU resources for training and inference, hindering their use in real-time or edge settings. Furthermore, efficient scaling in both model size and data remains a bottleneck, with challenges in data collection, curation, and model parallelization.

\subsection{Sample Efficiency}
\label{sec:sample}

Although progress has been made, particularly with meta-learning, contrastive objectives, and one/few-shot imitation, the demand for large volumes of video data remains a core challenge, especially for RL-based and high-capacity models. Many algorithms still require thousands of demonstrations or millions of interaction steps, which can be prohibitive in real-world settings. While approaches like feature extraction \ref{sec:feature_extraction}, goal-conditioned RL \ref{sec:rl_based}, and some hybrid models \ref{sec:hybrid_method} are relatively more data-efficient, the quest for robust learning from minimal or weakly-labeled data is ongoing. Bridging this gap is crucial for broadening the applicability of video-based robot learning in low-data regimes.

\subsection{Evaluation Metrics and Benchmarking}
\label{sec:metrics}

A critical challenge identified across the literature is the lack of standardized evaluation metrics and benchmarking protocols for video-based robot learning. Unlike computer vision or natural language processing, where large-scale public benchmarks drive progress, robotics evaluations are often fragmented and task-specific, relying on human judgment, custom setups, or bespoke datasets. This fragmentation complicates fair comparisons between approaches and slows the pace of reproducible progress. There is a growing need for community-driven, standardized benchmarks and clearly defined metrics that capture not only task success but also generalization, robustness, and sim-to-real performance.

\subsection{Causal Reasoning and Policy Abstraction}
\label{sec:causal}

A final major bottleneck, as surfaced in our comparative analysis, is the limited capacity of current methods to perform causal reasoning and high-level policy abstraction from video data. Most approaches focus on pattern recognition, goal inference, or direct imitation (see Section \ref{sec:approaches}), rarely incorporating causal structure or relational reasoning about actions and outcomes. As a result, models may lack robustness when faced with novel tasks or environments that require understanding of underlying cause-and-effect relationships. Bridging this gap will likely require new architectures and training paradigms that integrate causal inference, model-based reasoning, or neuro-symbolic methods, as well as datasets that explicitly capture causal interactions.
 
These challenges, distilled from a broad cross-section of the literature and subgroup analysis, highlight that progress in video-based robot learning is not uniform; each methodological paradigm offers distinct strengths but also faces recurring limitations. Addressing these bottlenecks, particularly in data transfer, scalability, and abstraction, will be key to achieving robust, generalizable, and efficient robot learning from videos in the future.

\section{Future Outlook}
\label{sec:future_outlook}
As discussed in previous sections, various approaches have been
proposed for learning manipulation skills through video
demonstrations. We also explored the challenges and limitations of
learning from videos. This section will spotlight several promising
but relatively underexplored areas in video-based learning
research. These areas include data efficiency, interactive and active
learning, multi-task learning architectures, integration of causal
reasoning, and the development of evaluation metrics and benchmarks.

\subsection{Tackling data efficiency and availability}
Addressing the challenges discussed in Section
\ref{sec:data_availability} regarding data availability and annotation
is crucial. Works such as
\cite{brohan2022rt,zitkovich2023rt,padalkar2023open,walke2023bridgedata,dasari2019robonet,fang2023rh20t}
have made dedicated efforts to collect extensive data for training
robots in various skills. While these endeavors contribute valuable
datasets, video demonstrations offer unique advantages compared to
task-specific datasets. Despite the risk of introducing biases, video
data provides a more unbiased and diverse representation of real-world
scenarios, fostering improved generalization. Additionally, videos
capture realistic dynamics and environmental variability, enabling
models to better handle uncertainties and variations encountered in
real-world scenarios.

As discussed in Section \ref{sec:data_availability}, poor
generalization may lead robots to struggle with tasks in states not
adequately covered during training. Generalization is a pervasive
topic in deep learning, and several works, including
\cite{brohan2022rt,chane2023learning,mandi2022towards}, propose
techniques for learning from videos to enhance model generalization
across diverse robots, tasks, and states. However, current approaches
still have limitations in their generalization, particularly to tasks
not recorded in the video demonstration data. Future work should focus
on addressing this limitation by identifying intuitive methods to
ensure not only generalization but also quick adaptation of these
models to changing tasks and environments.

\subsection{Improving data annotation through Active Learning}
Ensuring high-quality data for training models is crucial. Active
learning strategies, such as those proposed by \cite{1307137} and
\cite{settles2009active}, empower robots to strategically select
informative data points, optimizing the learning process by
Intelligently querying labels on challenging instances. This approach
reduces the reliance on extensive labeled datasets while maintaining
or improving performance.

Current approaches often passively observe large demonstration
datasets \cite{damen2018scaling,goyal2017something}, which can be
expensive to scale up. Integrating active physical trials on real
robots alongside video data observation combines the strengths of
imitation learning and embodied reinforcement learning. This approach
helps bridge the reality gap by incorporating interactions in physical
environments, dynamic feedback, and recalibration of visual
interpretations. It allows robots to adapt to environmental changes,
providing signals when contexts shift and offering opportunities for
adaptation. Passive video datasets often lack diversity across
potential deployment environments, making the inclusion of physical
interactions valuable. This approach not only enables the robot to
learn what works but also what doesn't work well in different
situations.

While studies like \cite{ramachandruni2020attentive} and
\cite{pathak2018zero} have proposed related techniques, they have not
thoroughly explored grounding policies learned from video data in
physical environments. Future works could benefit from addressing
these points.

\subsection{Tackling domain shift}
Future work should focus on addressing the persistent challenge of
domain shift between human and robot domains. Works like
\cite{wulfmeier2017addressing} addressed domain shift in the context
of appearance changes in outdoor robotics with adversarial domain
adaptation, while \cite{paudel2022learning} presented a survey in
learning for robot decision making under distribution shift. Advanced
domain adaptation techniques, including adversarial training and
meta-learning, could create more robust and generalizable
models. Multi-modal learning strategies that incorporate additional
sensory inputs may reduce reliance on visual domain
translation. Sim-to-real transfer methods and continual learning
paradigms offer promising avenues for improving domain
adaptation. Investigating attention mechanisms, unsupervised
techniques, and transformer-based architectures could yield more
effective domain-invariant features. Additionally, exploring causal
reasoning and few-shot learning approaches may enhance the efficiency
of skill transfer from human demonstrations to robotic
applications. By pursuing these strategies, future research can work
towards mitigating the impact of domain shift and improving the
effectiveness of learning from videos for robotic manipulation tasks.

\subsection{Introducing emerging techniques and architectures}
The techniques employed in the studies outlined in Section
\ref{sec:approaches} predominantly rely on single modalities or
involve single-task architectures. Recent research emphasizes the
efficacy of learning from multiple tasks and modalities
\cite{huang2021generalization,rahmatizadeh2018vision,devin2017learning,lee2019making}. Studies like \cite{mandi2022towards} and \cite{chane2023learning} discussed in Section \ref{sec:approaches} underscore the effectiveness of multi-task learning for acquiring robot manipulation skills from
videos. Challenges inherent in multi-tasking, extensively explored in studies like \cite{yu2020gradient}, become more pronounced due to the varying optimization constraints between predicting actions from fixed videos and closed-loop control problems.

Furthermore, the pursuit of learning manipulation skills from video has spurred the development of increasingly sophisticated generative architectures. These models have moved beyond simple regression or classification to generate complex, high-dimensional outputs like action trajectories and future video frames. The leading emerging architectural paradigms include diffusion models and world models.

Diffusion generative models have rapidly emerged as a dominant force in robotics, prized for their ability to model complex data distributions and their robustness in high-dimensional spaces \cite{wolf2025diffusion}. Their application to visuomotor control represents a significant step forward from prior generative approaches. Diffusion Probabilistic Models (DMs) operate on a simple yet powerful principle, executed in two stages \cite{wolf2025diffusion}. The first is a fixed forward process, where Gaussian noise is progressively added to a data sample (e.g., an image or an action trajectory) over a series of timesteps, gradually corrupting it into pure noise. The second stage is a learned reverse process, where a neural network, typically a U-Net or a Transformer, is trained to reverse this noising process. By learning to predict and remove the noise at each step, the model can start with a random noise tensor and iteratively denoise it to generate a new, high-fidelity sample from the original data distribution \cite{Parab2024DiffusionImitation}.

A key advantage of this framework for robotics is its inherent capacity to model multi-modal distributions \cite{wolf2025diffusion}. Many manipulation tasks do not have a single correct solution; there can be multiple valid trajectories to pick up an object or open a drawer. Traditional methods that predict a single, unimodal output (e.g., by minimizing mean squared error) tend to average these possibilities, resulting in mediocre or invalid actions. Diffusion models, by contrast, can capture the full distribution of successful behaviors, allowing them to generate diverse and plausible action sequences at inference time \cite{Parab2024DiffusionImitation}.

This capability has been elegantly harnessed in the "Diffusion Policies" framework for imitation learning \cite{Parab2024DiffusionImitation}. In this paradigm, the model learns a policy that generates robot actions directly from visual observations. The policy is a diffusion model trained to denoise an action trajectory conditioned on the current visual state of the environment. At each step of the reverse process, the model refines its prediction of the entire action sequence, leading to temporally coherent and precise behaviors \cite{ingelhag2024robotic}.

While diffusion policies excel at learning direct mappings from observations to actions, a parallel and complementary approach involves learning a predictive model of the world itself. These "world models" enable an agent to plan and reason by simulating the future consequences of its actions internally, a process often referred to as "imagining" \cite{Mendonca2023StructuredWM}. A world model learns the transition dynamics of an environment, formally represented as the probability distribution $p(s_{t+1}|s_t,a_t)$, where $s$ is the state and $a$ is the action. By repeatedly applying this learned model, an agent can forecast entire trajectories of future states that would result from a sequence of actions. This predictive capability is the foundation for model-based planning, where the agent can search for the optimal action sequence within its learned model before executing it in the real world. This approach can be significantly more sample-efficient than model-free methods, which must learn through extensive trial-and-error in the physical environment \cite{Yan2025TaskSpecificWorldModels}.

A powerful and intuitive instantiation of a world model is a video prediction model. Here, the state $s$ is represented by an image or a sequence of images, and the world model learns to generate future video frames conditioned on a sequence of actions. Mani-WM \cite{zhu2024maniwm} is a prime example of this approach, leveraging a diffusion transformer to generate high-resolution, long-horizon videos of a robot arm executing a specified action trajectory. The model employs a novel frame-level conditioning technique to ensure precise temporal alignment between the generated frames and the input actions. The resulting learned model serves as a high-fidelity, interactive simulator. This allows for downstream applications like policy evaluation and model-based planning to be conducted entirely with the generative model, mitigating the cost, safety concerns, and labor associated with extensive real-world robot rollouts \cite{zhu2024maniwm}.

\subsection{Robust evaluation metrics and benchmarks}
The rapid pace of architectural innovation necessitates equally rigorous and standardized evaluation methodologies. A model's claimed contributions are only as strong as the benchmarks used to validate them. We discussed below the most prominent benchmarks in video-based manipulation, analyzing their strengths, intended research focus, and, critically, their limitations, particularly concerning the evaluation of causal understanding.

\subsubsection{RLBench: A Testbed for Broad Skill Acquisition}
RLBench stands as a cornerstone for evaluating the breadth and generalization of manipulation skills \cite{9001253}. Its primary contribution is a massive and diverse suite of 100 unique, hand-designed tasks simulated in the V-REP (now CoppeliaSim) environment. These tasks span a wide spectrum of difficulty, from simple behaviors like reaching a target or opening a door to complex, multi-stage sequences like opening an oven and placing a tray inside.

A defining feature of RLBench is its rich multi-modal observation space, providing agents with RGB-D images from both a static over-the-shoulder camera and an eye-in-hand camera, as well as proprioceptive data like joint angles and torques. Perhaps its most unique and powerful feature is the provision of a virtually infinite supply of expert demonstrations for every task. These demonstrations are generated via motion planners operating on pre-defined waypoints, enabling a wide range of research in imitation learning and reinforcement learning that leverages expert data \cite{9001253}. The benchmark is explicitly designed to push research in multi-task learning, meta-learning, and, in particular, few-shot learning.

\subsubsection{CALVIN: The Standard for Long-Horizon Language Grounding}
While RLBench tests the breadth of skill acquisition, CALVIN (Composing Actions from Language and Vision) is the de facto standard for evaluating the depth of long-horizon, compositional reasoning \cite{9788026}. CALVIN is an open-source simulated benchmark, built in PyBullet, designed to develop and test agents that can solve complex manipulation tasks specified solely by natural language instructions. A single agent must learn to understand and execute a sequence of commands, such as "open the drawer, pick up the blue block, push the block into the drawer, open the sliding door" \cite{9788026}.

CALVIN's key contribution is its focus on long-horizon problems and language-based generalization. The benchmark includes four distinct environments (A, B, C, D) with shared structure but different visual textures and object layouts, allowing for rigorous testing of zero-shot generalization to novel scenes. The provided dataset is not a set of isolated, task-specific demonstrations, but rather hours of unstructured "play data," from which task sequences are procedurally labeled. This setup mimics a more realistic learning scenario where an agent must discover skills from continuous interaction data. Evaluation is performed on the agent's ability to generalize to novel language instructions and to complete long sequences of tasks, which is highly challenging as it requires the agent to robustly transition between different subgoals without compounding errors. Its status as a challenging and well-defined testbed has made it the proving ground for state-of-the-art models like VidMan \cite{wen2024vidman}.

\subsubsection{RoboTube: Bridging the Human-to-Robot Gap}
A central goal of the field is to enable robots to learn directly from observing humans. RoboTube is a benchmark designed specifically to facilitate research toward this goal \cite{pmlr-v205-xiong23a}. It directly addresses the limitations of prior video datasets, which often lack task complexity or relevance to household robotics. The RoboTube dataset consists of 5,000 high-quality, multi-view RGB-D video demonstrations of humans performing a variety of complex household tasks. These include the manipulation of not just rigid objects, but also articulated objects (drawers, cabinets), deformable objects (cloth), and granular materials (pouring).

The most significant feature of RoboTube is its "simulated twin" environment, RT-sim \cite{pmlr-v205-xiong23a}. The objects and scenes from the real-world videos have been meticulously 3D-scanned to create photo-realistic, physically accurate digital counterparts in simulation. This unique pairing of a real human video dataset with a high-fidelity simulated testbed is invaluable. It provides a controlled, reproducible platform for researchers to develop and benchmark algorithms for key challenges like sim-to-real transfer, representation learning from human video, and self-supervised reward learning, with the confidence that models validated in RT-sim have a clear path to deployment on a real robot. RoboTube aims to democratize research in this area by lowering the barrier to entry and providing a standardized platform for comparing different approaches to learning from human videos.

The design of a benchmark implicitly steers the research priorities of the community. The existence of CALVIN has catalyzed a wave of innovation in long-horizon, language-conditioned policies \cite{9788026}, while RLBench has standardized the evaluation of few-shot and multi-task learning \cite{9001253}. A critical analysis of these leading benchmarks, however, reveals a significant gap: none of them are explicitly designed to evaluate an agent's causal understanding of the world. An agent can achieve a high success rate on a CALVIN task by mastering the statistical correlations present in the massive demonstration dataset. It might learn that pushing a red button is followed by a light turning on, but it may not have learned the underlying causal link. This purely correlational policy would fail if the button's function were rewired, an object's physical properties were altered, or an unobserved confounder were introduced.

The MVP (Minimal Video Pairs) \cite{lopez2017discovering} benchmark, developed for visual question answering, offers a blueprint for how to address this gap. MVP consists of pairs of videos that are minimally different, often with a single changed detail, accompanied by the same question but with opposite answers. This design forces a model to move beyond superficial cues and engage in deeper reasoning to arrive at the correct answer. This principle must be extended to robotic manipulation to properly evaluate the benefits of causal models. A novel and impactful contribution would be the development of a new evaluation protocol, which could be termed "Causal-CALVIN" or "Interventional RLBench." In this protocol, an agent would first be trained on the standard benchmark dataset. Then, its generalization and robustness would be tested on a suite of evaluation tasks where the underlying causal structure of the environment has been perturbed. For example, the mass or friction of an object could be significantly changed, the causal link between a switch and a light could be broken or rewired to a different switch, or a previously free-sliding drawer could be made to stick. A purely correlational model, having overfitted to the statistics of the training environment, would be expected to fail catastrophically. In contrast, a model equipped with an accurate causal model of the world should be able to either adapt its policy to the new dynamics or at least recognize that its model of the world is no longer valid, enabling more robust and intelligent failure recovery. This provides a concrete, quantitative methodology for measuring the tangible benefits of causal reasoning, moving evaluation beyond simple task success rates to a more meaningful assessment of physical understanding.

Table \ref{tab:benchmark_comparison} provides a strategic overview of these benchmarks, enabling the selection of appropriate evaluation platforms to highlight different facets of a proposed model's performance.

\begin{table*}[h!]
\centering
\caption{A Comparison of Benchmarks for Robotics}
\label{tab:benchmark_comparison}
% \begin{tabularx}{\textwidth}{@{} l X c X X l X @{}}
\begin{tabularx}{\textwidth}{@{} p{1.1cm} p{1.8cm} p{1.2cm} p{2.2cm} p{2.0cm} p{1.6cm} p{3.0cm} @{}}
\toprule
\textbf{Benchmark} & \textbf{Primary Focus} & \textbf{\# of Tasks} & \textbf{Data Provided} & \textbf{Key Evaluation Metric(s)} & \textbf{Simulation Environment} & \textbf{Suitability for Causal Evaluation} \\
\midrule

CALVIN & Long-horizon, language-conditioned, compositional tasks. & 34 & \textasciitilde{24} hours of unstructured, teleoperated "play" data; language annotations for \textasciitilde{1}\% of data. & Long-Horizon Multi-Task Language Control (LH-MTLC) success rate. & PyBullet & Low (as is); High (with modification). The current setup rewards correlational learning. A "Causal-CALVIN" variant with perturbed physics/mechanisms would be required to test causal understanding. \\
\midrule

RLBench & Few-shot, meta-, and multi-task learning; broad skill acquisition. & 100 & Infinite supply of motion-planned expert demonstrations for every task; multi-modal observations (RGB-D, proprioception). & Few-shot task success rate. & CoppeliaSim (V-REP) & Low (as is); High (with modification). The diversity of tasks is high, but the physics within each task is fixed. An "Interventional RLBench" with variations in physical properties (mass, friction, etc.) would be needed. \\
\midrule

RoboTube & Learning from human video demonstrations; human-to-robot transfer. & \textasciitilde{50} & 5,000 multi-view RGB-D videos of human demos; a "simulated twin" (RT-sim) with photo-realistic assets. & Policy success rate in RT-sim after training on human videos; sim-to-real transfer success. & Custom (RT-sim) & Moderate. The paired real/sim setup is ideal for studying the transfer of causal models. The diversity of object types (deformable, granular) provides a rich testbed for models of complex causal interactions. \\

\bottomrule
\end{tabularx}
\end{table*}

\subsection{Integration of causal reasoning}
To build robots that can operate robustly and adaptively in the open world, it is essential to move beyond learning statistical correlations and toward models that capture the underlying causal structure of their environment. Consequently, below we provide a detailed investigation into the principles and methods of causal reasoning as they apply to video-based manipulation, establishing the foundation for a research agenda centered on this critical capability.

The vast majority of modern machine learning models, including deep neural networks, are powerful function approximators trained to minimize a loss function on an independently and identically distributed (i.i.d.) dataset. This process enables them to excel at learning complex correlations within the training data. However, correlation does not imply causation. This fundamental limitation is the root cause of their brittleness when deployed in the real world, which is inherently non-stationary and subject to constant distribution shifts \cite{Lee-2024-141571}. A robot trained in a lab may learn a spurious correlation between the color of a block and its weight, and will fail when presented with a block of a different color.

Causal models offer a principled escape from this trap \cite{li2020causal}. By aiming to represent the actual data-generating processes, causal models provide a foundation for true generalization. A causal model understands that an object's mass, not its color, determines the force required to lift it. This understanding allows for robust performance even when encountering novel objects and conditions. For robotics, the promise of causality is threefold: robustness to environmental changes, generalization to novel scenarios, and the ability to perform counterfactual reasoning (e.g., "what would have happened if I had pushed the object instead of grasping it?"). Which is the bedrock of intelligent planning and decision-making \cite{li2020causal}.

\subsubsection{Methods for Causal Discovery from Visual Data}
Causal discovery is the process of inferring the causal structure (typically represented as a Directed Acyclic Graph, or DAG) from data \cite{li2023deep}. In robotics, this means discovering the cause-and-effect relationships between objects, actions, and environmental variables from sensory inputs like video. These methods can be broadly categorized into observational and interventional approaches.

\begin{itemize}
    \item \textbf{Observational Approaches:} Observational methods attempt to uncover causal structure from passively collected data, without actively manipulating the system. These approaches typically fall into two families: constraint-based and score-based methods. Constraint-based algorithms, like the PC algorithm \cite{li2023deep}, work by performing a series of conditional independence tests on the data to prune edges from a fully connected graph. Score-based methods define a scoring function (e.g., Bayesian Information Criterion) that measures how well a given graph structure fits the data and then search the space of possible graphs for the one with the best score.

    Applying these methods to high-dimensional video data requires specialized architectures. The Visual Causal Discovery Network (V-CDN) \cite{li2020causal} is a seminal work in this area. V-CDN is an end-to-end model that learns to discover causal relationships in physical systems directly from video. It consists of three key modules:

    \begin{enumerate}
        \item Perception Module: Extracts an unsupervised, temporally consistent keypoint representation of objects in the scene from raw images. These key points serve as the variables in the causal graph.
        
        \item Inference Module: Observes the dynamics of these keypoints over a short video sequence and infers a latent causal graph, determining which keypoints are causally related (e.g., connected by a spring or a rigid rod).
        
        \item Dynamics Module: A graph neural network that takes the inferred causal graph as input and learns to predict the future evolution of the system.
    \end{enumerate}

    A crucial aspect of V-CDN is its assumption that the training data, while passively observed, is sourced from a variety of configurations and environmental conditions. This is treated as data from "unknown interventions" on the system \cite{li2020causal}. For example, by observing videos of systems with different numbers of objects or different spring constants, the model can disambiguate direct causal links from mere correlations and identify the correct underlying causal graph without requiring explicit labels for the interventions performed \cite{li2020causal}.

    \item \textbf{Interventional Approaches:} While observational methods are powerful, the gold standard for establishing causality is intervention: the act of actively manipulating a variable, denoted as $do(X=x)$, and observing the effect on the system \cite{pmlr-v186-mendez-molina22a}. Interventions break potential confounding pathways and provide unambiguous evidence of cause-and-effect relationships. Robots, as physically embodied agents, are uniquely positioned to perform such interventions, making them ideal platforms for active causal discovery \cite{castri2025causality}.

    A powerful demonstration of this principle is SCALE (Skills from CAusal LEarning) \cite{pmlr-v229-lee23b}. SCALE addresses the problem of learning manipulation skills that generalize across different contexts. Instead of learning a single, complex policy over a high-dimensional state space, SCALE uses a simulator as a "causal reasoning engine" to perform targeted interventions. For a given task, it systematically perturbs context variables (e.g., object positions, sizes, masses) and observes whether the intervention affects the task outcome. This process allows it to identify the minimal subset of context variables that are causally relevant for success. It then learns a "compressed" skill or option that is conditioned only on this small set of causal variables, ignoring all spurious features. This results in policies that are dramatically more sample-efficient and exhibit superior sim-to-real transfer, as they are not distracted by irrelevant, correlational features of the training environment \cite{pmlr-v229-lee23b}.

    Building on this, the Causal Robot Discovery (CRD) framework proposes a continual, online approach to causal learning \cite{pmlr-v208-castri23a}. The robot begins by building an initial causal model from passive observation. It then analyzes this model to identify the most uncertain or unreliable links (e.g., those with high p-values from conditional independence tests). Based on this uncertainty, the robot plans and executes its next set of interventions specifically to gather data that will maximally resolve this uncertainty \cite{pmlr-v208-castri23a}. This creates an efficient, self-improving feedback loop where the robot actively seeks the most informative data to refine its causal understanding of the world, making it particularly well-suited for resource-constrained robotics applications \cite{pmlr-v208-castri23a}.
    
\end{itemize}

\subsubsection{Causal Representation Learning (CRL)}
The effectiveness of any causal discovery method depends on the variables over which it operates. Causal Representation Learning (CRL) is an emerging field that aims to bridge the gap between low-level sensory data (like pixels) and high-level causal variables \cite{CVPR2024CorrWorkshop}. The goal of CRL is to learn a mapping from high-dimensional observations to a low-dimensional latent space where the axes of the representation correspond to the independent causal mechanisms of the world \cite{Lee-2024-141571}. For example, an ideal causal representation of a scene would disentangle factors like object identity, pose, lighting, and background into separate, independently controllable latent variables. Such a representation is inherently more compositional and generalizable, as the model can reason about and manipulate these factors independently, a key requirement for advanced robotics and embodied AI \cite{CVPR2024CorrWorkshop}.

The motivation for incorporating causality into robotics must be driven by tangible, pragmatic benefits that improve manipulation performance. It is not merely a pursuit of philosophical purity or interpretability. The evidence from recent work clearly demonstrates that causal reasoning is a practical tool for building more efficient, robust, and intelligent robots.

The most direct evidence comes from the SCALE framework, which links causal discovery directly to improved policy learning \cite{pmlr-v229-lee23b}. By using interventions in a simulator to identify the true causal drivers of task success, SCALE learns a compressed policy that is conditioned only on relevant variables. This policy is not only more sample-efficient to train but also more robust to spurious correlations in the environment. When transferred to the real world, this causally-informed policy succeeds where a standard, correlational policy fails, providing a clear demonstration of how causal feature selection enhances generalization and sim-to-real transfer.

Furthermore, causal knowledge can make the entire learning process more efficient. A robot that understands the causal structure of its environment can guide its exploration more intelligently. Instead of exploring randomly, an RL agent can use its causal model to prioritize actions that are most likely to influence task-relevant variables, dramatically reducing the number of samples needed to learn an effective policy \cite{pmlr-v186-mendez-molina22a}. The CRD framework operationalizes this by using the uncertainty in its current causal model to actively plan the most informative interventions, creating a highly efficient data collection loop \cite{pmlr-v208-castri23a}.

Finally, a validated causal model unlocks the ability to perform counterfactual reasoning, which is the foundation of robust, deliberative planning. The model can simulate the outcomes of actions it has never taken in situations it has never seen, allowing it to plan for novel circumstances and recover from failures \cite{li2020causal}. For instance, it can be used to analyze why a task failed by tracing back the chain of causal events that led to the undesirable outcome, enabling more sophisticated error diagnosis and correction \cite{pmlr-v229-lee23b}.

Table \ref{tab:causal_methods} provides a taxonomy of key causal discovery methods relevant to robotics, organizing the literature into a coherent framework and highlighting their practical applications.

\begin{table*}[h!]
\centering
\caption{Comparison of Causal Discovery Methods in Robotics}
\label{tab:causal_methods}
\begin{tabularx}{\textwidth}{@{} l X X X c @{}}
\toprule
\textbf{Method} & \textbf{Core Principle} & \textbf{Data Requirement} & \textbf{Key Application/Benefit} & \textbf{Citation} \\
\midrule

V-CDN & End-to-end discovery of latent causal graphs from video via keypoint extraction and graph inference. & Observational (passive videos from diverse, "unknown" interventional settings). & Enables learning of predictive dynamics models that can perform counterfactual reasoning and extrapolate to unseen system configurations. & 9 \\
\midrule

SCALE & Active interventional discovery of causally relevant features for a task to learn a compressed, robust skill. & Interventional (requires a simulator or "causal reasoning engine" to perform targeted interventions). & Improves sample efficiency and sim-to-real transfer by learning policies that are robust to spurious correlations. & 11 \\
\midrule

CRD & Continual, online refinement of a causal model by using model uncertainty to guide the next set of active interventions. & Hybrid (Observational + Interventional). The robot actively collects interventional data to improve its model. & Enables efficient causal discovery on resource-constrained robots by creating a self-improving, active learning loop. & 43 \\
\midrule

F-PCMCI & An efficient, filtered version of the PCMCI algorithm for causal discovery from time-series data. & Observational (time-series data). & Designed for fast and accurate causal analysis in real-time robotics applications, such as modeling human-robot interaction. & 45 \\

\bottomrule
\end{tabularx}
\end{table*}

\section{Conclusion}
\label{sec:conclusion}
In this survey, we reviewed the emerging paradigm of robot learning for manipulation skills by leveraging abundantly available uncurated videos. Learning from video data allows for better generalization, reduction in dataset bias, and cutting down the costs associated with obtaining well-curated datasets. We began by outlining and discussing the essential components required for learning from video data and some current large-scale datasets and network architectures proposed for robot learning.

We surveyed techniques spanning representational, reinforcement, imitation, hybrid, and multimodal learning approaches for learning from demonstration videos in an end-to-end or modular manner. Analysis was provided around representations, sample efficiency, interpretability, and robustness for categories like pose estimation, image translation, and vision-language
approaches. The benefits highlighted include generalization beyond controlled
environments, scalability through abundant supervision, and avoiding biases coupled with human dataset curation. We also discussed evaluation protocols, sim-to-real challenges, and interactive learning as augmentations to pure video-based learning.

In conclusion, while still a nascent research direction, robot learning from online human videos shows immense promise in overcoming key data challenges prevalent in other supervised manipulation learning paradigms. If open challenges around dynamics, long-horizon understanding, absence of consistent and objective evaluation and benchmarking protocols, and sim-to-real transfer are systematically addressed, video-based learning can provide a scalable, economical, and generalizable pathway for robot acquisition of intricate real-world manipulation skills.

\vskip 0.2in
\bibliographystyle{IEEEtran}
\bibliography{IEEEabrv,main.bib}

\begin{IEEEbiography}[{\includegraphics[width=1in,height=1.25in,clip,keepaspectratio]{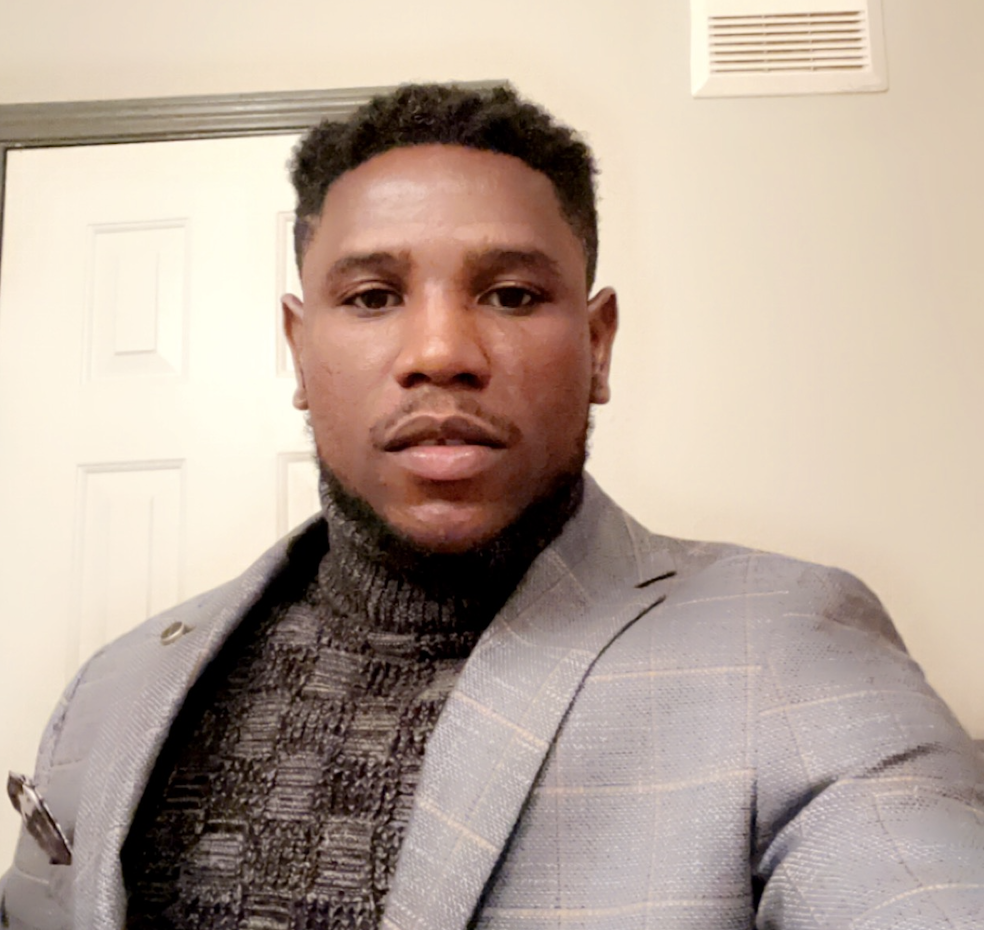}}]{Chrisantus Eze} received the B.Eng. degree in Electrical and Electronic Engineering from the Federal University of Technology, Owerri in Nigeria and is currently pursuing a Ph.D. in computer science from Oklahoma State University.

From 2019 to 2021, he was a Software Engineer at Seamfix Limited, in Lagos Nigeria. Since January 2022, he has been a PhD student at the
Computer Science Department, Oklahoma State University, Stillwater.
He is the author of three conference publications. His research focuses on improving long-horizon and spatial reasoning in robots for manipulation. To achieve this, he proposes and implores techniques, algorithms and systems for computer vision, foundation models, and imitation learning.
\end{IEEEbiography}

\begin{IEEEbiography}[{\includegraphics[width=1in,height=1.25in,clip,keepaspectratio]{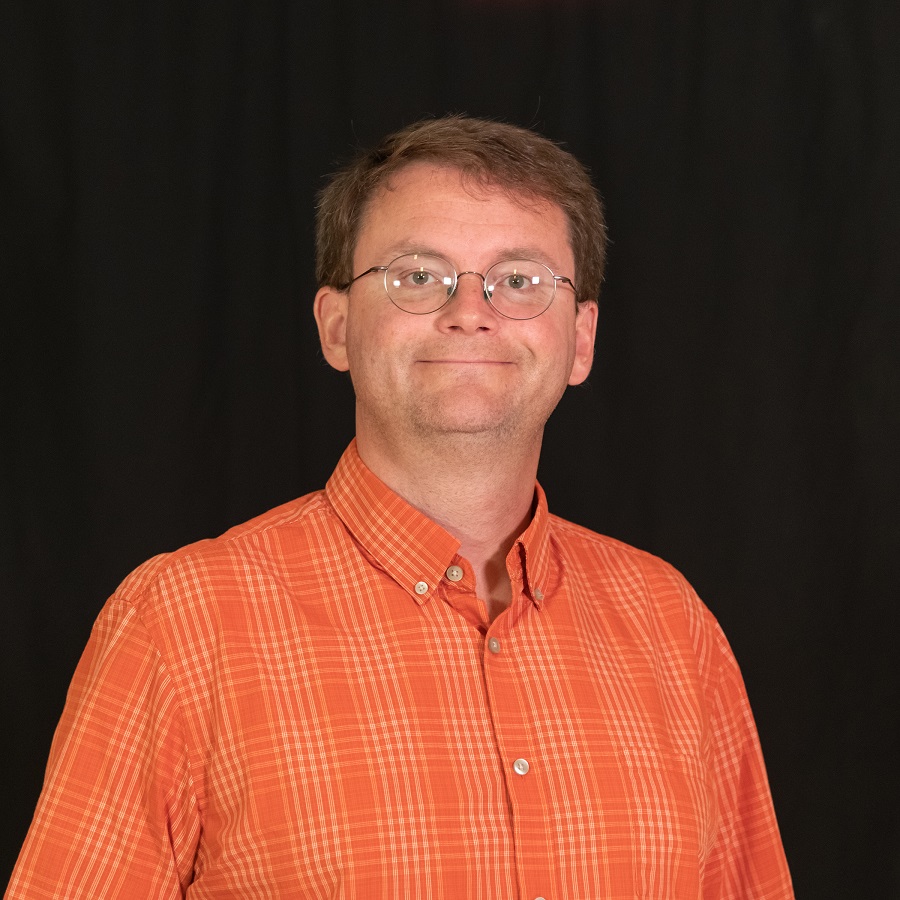}}]{Christopher Crick} Professor Christopher Crick received the B. A in History and an M.S. in Computer Science from Harvard University and a PhD in Computer Science from Yale University in 2009 after which he joined Brown University as a postdoctoral fellow.

He joined Oklahoma State University as a Professor in 2018 and carries out research at the intersection of robotics and explainable AI. His research agenda is motivated by two overarching, complementary goals: to ground models of developmental psychology and cognitive science in realized, embodied computational systems; and to use models of human cognition and social development to improve robot control and learning, human-robot interaction, and artificial intelligence in general. 

\end{IEEEbiography}

\EOD

\end{document}